\ifdefined\pdftexversion\pdfoutput=1\fi
\documentclass{article}

\PassOptionsToPackage{numbers}{natbib}
\usepackage[preprint]{neurips_2026}
\usepackage{titlesec}
\usepackage{titletoc}        
\usepackage{caption}
\titlespacing*{\section}      {0pt}{6pt plus 1pt minus 1pt}{3pt}

\usepackage[utf8]{inputenc} 
\usepackage[T1]{fontenc}    
\usepackage{hyperref}       
\hypersetup{
		colorlinks=true,
		linkcolor=blue!70!black,
		citecolor=blue!70!black,
		urlcolor=blue!70!black,
		pdftitle={OpenThoughts-Agent: Data Recipes for Agentic Models},
		pdfauthor={The OpenThoughts Team}}
\usepackage{url}            
\usepackage{booktabs}       
\usepackage{amsfonts}       
\usepackage{nicefrac}       
\usepackage{microtype}      
\usepackage{xcolor}         
\usepackage{graphicx}       
\graphicspath{{figures/}{../figures/}}
\makeatletter
\providecommand\input@path{}
\g@addto@macro\input@path{{tables/}{../tables/}}
\makeatother
\usepackage{placeins}
\usepackage{booktabs, multirow, xcolor, colortbl, array, rotating, graphicx, caption}
\usepackage{caption}
\usepackage{amsmath}
\usepackage{multirow}
\usepackage{rotating}
\usepackage{cleveref}
\usepackage{todonotes}
\usepackage{enumitem}

\title{\centering
  \raisebox{-0.15cm}{\includegraphics[scale=.16]{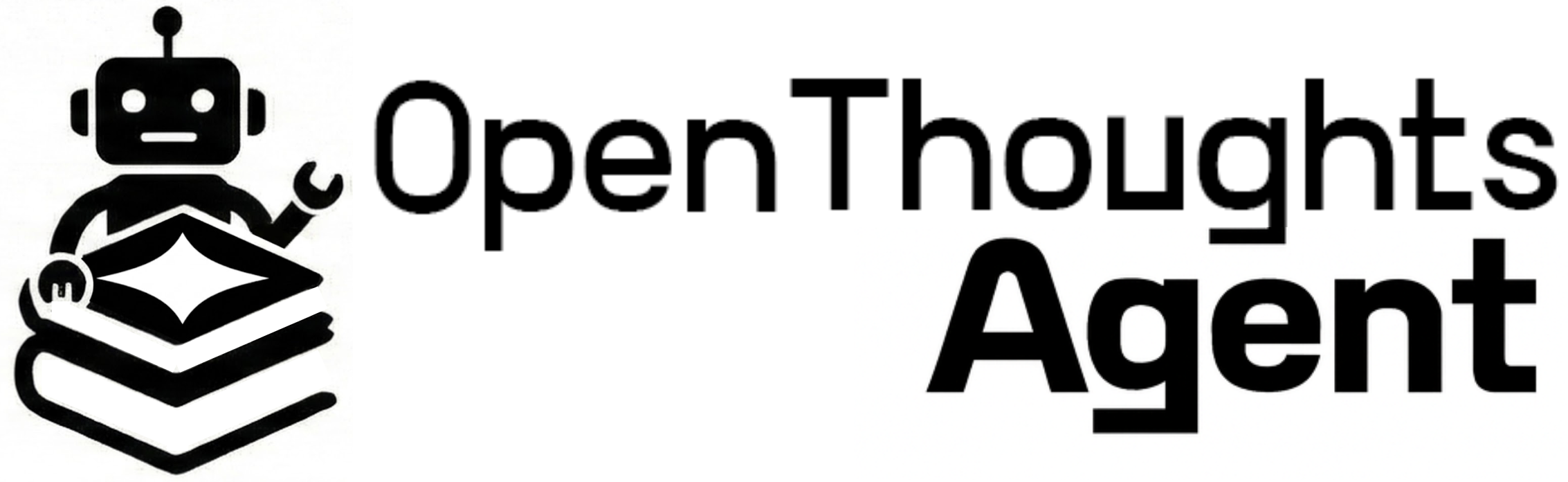}} \\ Data Recipes for Agentic Models
}
\author{%
{\normalfont
\parbox{0.95\textwidth}{\centering
\vspace{0.2cm}
Negin Raoof$^{*1}$,
Richard Zhuang$^{*2}$,
Marianna Nezhurina$^{*3,4,5}$,
Etash Guha$^{*2}$, \\
Atula Tejaswi$^{6}$,
Ryan Marten$^{7}$,
Charlie F. Ruan$^{1}$,
Tyler Griggs$^{1}$,
Alexander Glenn Shaw$^{8}$,
Hritik Bansal$^{9}$,
E. Kelly Buchanan$^{2}$,
Artem Gazizov$^{10}$,
Reinhard Heckel$^{11}$,
Chinmay Hegde$^{12}$,
Sankalp Jajee$^{13}$,
Daanish Khazi$^{14}$,
Emmanouil Koukoumidis$^{15}$,
Xiangyi Li$^{16}$,
Hange Liu$^{17}$,
Shlok Natarajan$^{2}$,
Harsh Raj$^{18}$,
Nicholas Roberts$^{19}$,
Ethan Shen$^{20}$,
Nishad Singhi$^{21}$,
Michael Siu$^{22}$,
Ashima Suvarna$^{9}$,
Hanwen Xing$^{22}$,
Patrick Yubeaton$^{12}$,
Robert Zhang$^{6}$,
Leon Liangyu Chen$^{2}$,
Xiaokun Chen$^{2}$,
Steven Dillmann$^{2}$,
Saadia Gabriel$^{9}$,
Xunyi Jiang$^{23}$,
Anurag Kashyap$^{24}$,
Boxuan Li$^{25}$,
Yein Park$^{26}$,
Minh Pham$^{12}$,
Sujay Sanghavi$^{6}$,
Lin Shi$^{27}$,
Ke Sun$^{17}$,
Yixin Wang$^{28}$,
Zhiwei Xu$^{28}$,
Erica Zhang$^{2}$,
Siyan Zhao$^{9}$,
Wanjia Zhao$^{2}$,
Jenia Jitsev$^{3,4,5}$,
Alex Dimakis$^{1,7}$, \\
Benjamin Feuer$^{\dagger 2,15}$,
Ludwig Schmidt$^{\dagger 2}$
\\[0.7em]
{\small
$^{1}$UC Berkeley,
$^{2}$Stanford University,
$^{3}$JSC,
$^{4}$LAION,
$^{5}$Open-$\Psi$ (Open-Sci) Collective,
$^{6}$University of Texas at Austin,
$^{7}$Bespoke Labs,
$^{8}$Laude Institute,
$^{9}$UCLA,
$^{10}$Harvard University \& Harvard Medical School,
$^{11}$TU Munich \& Munich Center for Machine Learning,
$^{12}$New York University,
$^{13}$Medical University of South Carolina,
$^{14}$The LLM Data Company,
$^{15}$Oumi.AI,
$^{16}$BenchFlow,
$^{17}$Independent Researcher,
$^{18}$Northeastern University,
$^{19}$University of Wisconsin--Madison,
$^{20}$University of Washington,
$^{21}$TU Darmstadt,
$^{22}$University of Southern California,
$^{23}$UC San Diego,
$^{24}$Amazon,
$^{25}$Microsoft,
$^{26}$Korea University,
$^{27}$Cornell Tech,
$^{28}$University of Michigan}
\\[0.5em]
{\footnotesize $^{*}$Equal contribution.\quad $^{\dagger}$Equal contribution.}
}}%
}

\begin{document}

\maketitle

\begin{abstract}
Agentic language models dramatically expand the applications of AI yet little is publicly known about how to curate training data for broadly capable agents.
Existing open efforts such as SWE-Smith, SERA, and Nemotron-Terminal typically target a single benchmark, leaving open the question of how to train models that generalize across diverse agentic tasks.
The OpenThoughts-Agent (OT-Agent) project addresses this gap with a fully open data curation pipeline for training agentic models.
We conduct more than 100 controlled ablation experiments to systematically investigate each stage of the pipeline, yielding insights on the importance of task sources and diversity.
We then assemble a training set of 100K examples from our pipeline and fine-tune Qwen3-32B on this dataset, which yields an average accuracy of 44.8\% across seven agentic benchmarks and a 3.9 percentage point improvement over the strongest existing open data agentic model (Nemotron-Terminal-32B, 40.9\%).
Moreover, our training data exhibits strong scaling properties, outperforming alternative open datasets at every training set size in compute-controlled comparisons. We publicly release our training sets, data pipeline, experimental data, and models at openthoughts.ai to support future open research on agentic model training.

\end{abstract}

\pagebreak

\begin{figure*}[t]
  \centering
  \includegraphics[width=\textwidth]{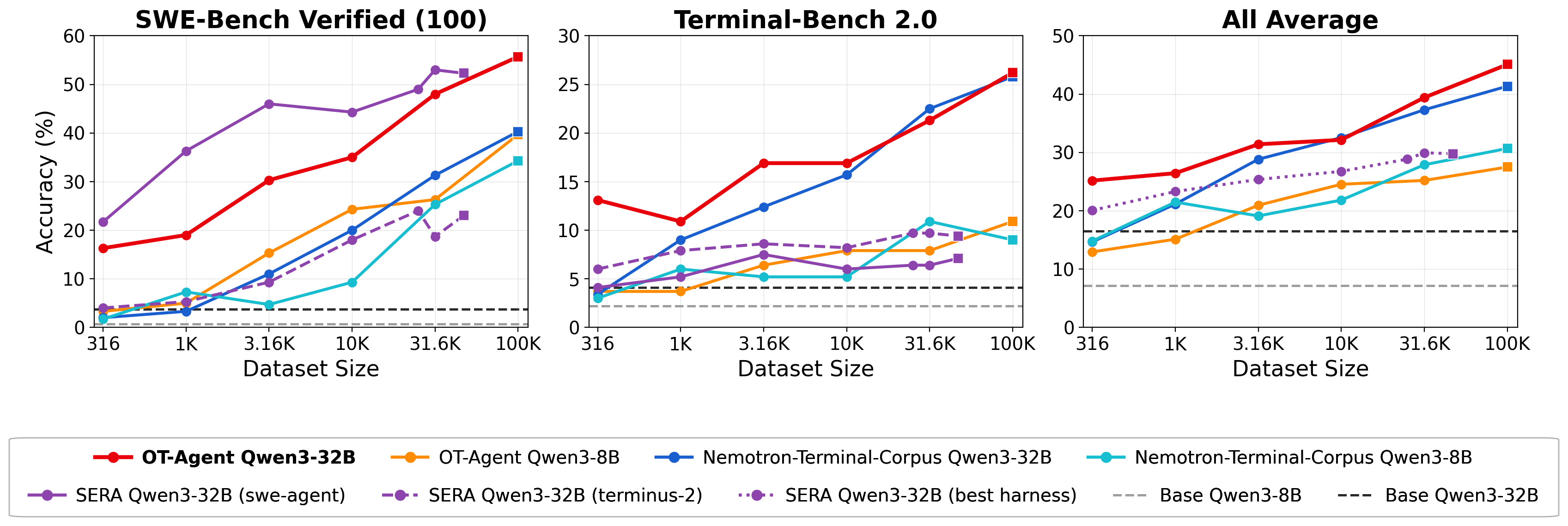}
  \caption{\textbf{The OpenThoughts-Agent-SFT dataset leads to SotA performance on
Terminal-Bench 2.0 and an 100-subset of SWE-Bench Verified at large dataset scales.}
The all-benchmark average is over the seven benchmarks reported in
Table~\ref{tab:ot_agent_main_table1}. Additional notes on SERA below.\protect\footnotemark}
  \label{fig:scaling-curves}
\end{figure*}
\footnotetext{SERA uses SWE-agent for data generation and evaluation, so we report both the SWE-agent (solid) and Terminus-2 (dashed) harnesses on SWE-Bench and Terminal-Bench, and the best of the two harnesses per benchmark for the average. SERA's dataset contains at most 47K examples, so its scaling curve stops at 47K.}

\section{Introduction}
\label{sec:intro}

Agentic language models have dramatically expanded the applications of AI.
Instead of only answering questions, this new generation of models can perform a wide range of complex tasks that involve using a computer in intricate ways.
As a result, AI agents such as Claude Code, Codex, and OpenClaw have rapidly grown in popularity.
To enable these agentic applications, the underlying models have improved in their ability to use tools and reason coherently over a long horizon.
Further developing these models towards better agents is one of the most important research directions in AI.

At the same time, the public literature contains little information on how to train state-of-the-art agentic models, especially when it comes to training data.
The recent DeepSeekV4 release is an illustrative example~\citep{dsv4}: the model weights are open and the accompanying paper contains more than 50 pages with details on the architecture and training process, but the paper describes the training data only at a high level with two paragraphs.
While there are initial efforts to curate training data for agents in the open such as SWE-Smith \citep{yang2025swesmithscalingdatasoftware}, SERA \citep{shen2026serasoftverifiedefficientrepository}, Nemotron-Terminal \citep{pi2026dataengineeringscalingllm}, and OpenSWE \citep{fu2026davincienvopensweenvironment}, these efforts usually focus only on one benchmark at a time, e.g., SWE-Bench~\citep{jimenez2024swebenchlanguagemodelsresolve} or Terminal-Bench~\citep{merrill2026terminalbenchbenchmarkingagentshard}.
Hence, it is currently difficult for open AI research to understand how broadly capable agentic models are trained, and to contribute to their improvement.

We take a step towards open training data for agentic models with the OpenThoughts-Agent (OT-Agent) project.
Building on the insights from the prior OpenThoughts work \citep{guha2025openthoughtsdatarecipesreasoning}, we focus on post-training data for supervised fine-tuning (SFT), now with the goal of improving a model’s performance on multiple agentic benchmarks.
Our first contribution is a comprehensive data curation pipeline for agentic SFT data on which we conduct more than 100 ablation experiments.
Our experiments lead to the following key findings:
\begin{itemize}[topsep=0pt, itemsep=2pt, parsep=0pt, leftmargin=5.5mm]
\item As with reasoning data, the choice of instructions is among the most important factors in our data pipeline.
\item The strongest model by benchmark performance does not necessarily make the best teacher.
\item Filtering training data to retain the execution traces with more model turns improves the resulting training sets.
\item Repeating the top few sources leads to diminishing returns in our largest training runs, and we therefore expand the set of data sources to increase diversity.
\end{itemize}

\begin{table}[t!]
\centering
\newcolumntype{C}[1]{>{\centering\arraybackslash}p{#1}}
\resizebox{\textwidth}{!}{
\begin{tabular}{p{0.15cm} l | C{1.0cm}C{1.0cm}C{1.0cm}C{1.0cm}C{1.0cm}C{1.0cm} @{\hspace{0.15cm}}p{0.05cm}@{\hspace{0cm}} C{1.0cm}}
\toprule
 & Benchmark & \rotatebox{45}{\textbf{OpenThinkerAgent-32B}} & \rotatebox{45}{\textbf{Nemotron-Terminal-32B}} & \rotatebox{45}{\textbf{SWE-Lego-Qwen3-32B}} & \rotatebox{45}{\textbf{SERA-32B}} & \rotatebox{45}{\textbf{SA-SWE-32B}} & \rotatebox{45}{\textbf{DeepSWE-Preview}} & \rotatebox{45}{\scriptsize{------ Base Model ------}} & \rotatebox{45}{\textbf{Qwen3-32B}} \\
\midrule
 & Train Size & {100K} & {264K} & {18K} & {25K} & {4.5K} & {4.5K} & \multicolumn{1}{!{\vrule width 0.5pt}c}{} & {N/A} \\
 & Method & {SFT} & {SFT} & {SFT} & {SFT} & {RL} & {RL} & \multicolumn{1}{!{\vrule width 0.5pt}c}{} & {N/A} \\
\midrule
\multirow{1}{*}{} & Average (core and OOD evals) & \textbf{44.8} & 40.9 & 34.7 & 28.1 & 26.9 & 26.7 & \multicolumn{1}{!{\vrule width 0.5pt}c}{} & 22.8 \\
\midrule
\multirow{2}{*}{\rotatebox{90}{\textit{Core}}} & SWE-Bench-Verified & \textbf{54.0} & 41.9 & 51.0 & 49.4 & 39.4 & 42.2 & \multicolumn{1}{!{\vrule width 0.5pt}c}{} & 29.1 \\
 & Terminal-Bench 2.0 & \textbf{26.2} & \textbf{25.1} & 16.1 & 9.7 & 16.2 & 4.9 & \multicolumn{1}{!{\vrule width 0.5pt}c}{} & 7.5 \\
\midrule
\multirow{5}{*}{\rotatebox{90}{\textit{OOD}}} & Aider-Polyglot & \textbf{32.4} & 24.9 & 30.1 & 26.7 & 17.3 & 27.3 & \multicolumn{1}{!{\vrule width 0.5pt}c}{} & 28.9 \\
 & BFCL-Parity & \textbf{85.9} & 69.1 & 81.0 & 69.1 & 74.8 & 77.2 & \multicolumn{1}{!{\vrule width 0.5pt}c}{} & 68.3 \\
 & MedAgentBench & 47.8 & \textbf{62.6} & 36.2 & 15.6 & 15.8 & 8.7 & \multicolumn{1}{!{\vrule width 0.5pt}c}{} & 6.8 \\
 & GAIA-127 & \textbf{23.6} & \textbf{22.3} & 12.9 & 8.7 & 11.5 & 16.5 & \multicolumn{1}{!{\vrule width 0.5pt}c}{} & 9.7 \\
 & FinanceAgent-Terminal & \textbf{44.0} & 40.7 & 15.3 & 17.3 & 13.3 & 10.0 & \multicolumn{1}{!{\vrule width 0.5pt}c}{} & 9.3 \\
\bottomrule
\end{tabular}}
\vspace{6pt}
\caption{\textbf{OpenThinkerAgent-32B is the best open-data model (Qwen-3 model family or earlier, up to 32B scale) on an average of seven agentic benchmarks.} Our model is the best overall on SWE-Bench-Verified and Terminal-Bench 2.0, and also generalizes well on OOD benchmarks we did not use during development. We evaluate each model both in Terminus-2 and in the model's original harness and, for each model, report the maximum accuracy over both harnesses. All models in this table are trained from Qwen3-32B. Bolded cells are within one standard error of the maximum in each row; more detail about standard error can be found in Appendix Table~\ref{tab:appendix_eval_table}.}
\label{tab:ot_agent_main_table1}
\end{table}

Building on our experiments, we assemble a state-of-the-art training set for fine-tuning agentic models.
In particular, we fine-tune the Qwen3-32B model on 100k data points from our pipeline and achieve the best performance on a broad suite of agentic tasks compared to other open data models with a Qwen3 or earlier base model at <=32B scale (see Table \ref{tab:ot_agent_main_table1}).
Our model achieves 54.0\% on SWE-Bench Verified and 26.2\% on Terminal-Bench 2.0, compared to 41.9\% and 25.1\% for Nemotron-Terminal-32B.
In addition, our model also outperforms prior work on further agentic benchmarks including Aider Polyglot \citep{aiderpolyglot2024}, BFCL-Parity \citep{patil2025bfcl}, GAIA-127 \citep{mialon2023gaia}, and FinanceAgent-Terminal \citep{valsai2024financeagent}. 
Figure \ref{fig:scaling-curves} shows that our training set is not only good because of its size but also exhibits strong scaling trends, outperforming other open datasets in a compute-controlled way for every training set size.

In addition to our SFT investigation, we also study data curation for reinforcement learning (RL).
We briefly document the challenges of existing open-data RL curation efforts (such as reproducibility, usability, and scaling) and introduce a newly curated RL dataset. We then validate the efficacy of this data by post-training an 8B model in two stages (SFT + RL) which outperforms our best single-stage 8B model, as well as the strongest existing <=8B baselines, on average across 7 agentic benchmarks (see Table \ref{tab:ot_agent_8B_table3} for details).

We publicly release our training sets, data pipeline, experimental data, and models at \url{https://www.openthoughts.ai/}, so that future open research can build on our artifacts.

\looseness=-1

\vspace{-0.25cm}
\section{Related Work}
\label{sec:related}

\textbf{Data curation.} Data curation includes data sourcing, data labeling / verification, and sometimes filtration~\citep{gadre2023datacompsearchgenerationmultimodal,li2025datacomplmsearchgenerationtraining}. The advent of large language models and scaling laws progressively made clear the large role data curation plays in AI capabilities, and led to considerable industrial research and rapid advances in the state-of-the-art, as well as a series of public data curation efforts for vision-language models, language models and reasoning models~\citep{gadre2023datacompsearchgenerationmultimodal,li2025datacomplmsearchgenerationtraining,guha2025openthoughtsdatarecipesreasoning,bercovich2025llamanemotronefficientreasoningmodels}. Rigorous public research on data curation for agents, however, remains scarce, providing a core motivation for this work.

\textbf{Agents and their benchmarks.} The evaluation of large language models has progressed rapidly in the space of the last five years; where pioneering models such as GPT-3 reported primarily on multiple-choice benchmarks such as MMLU, evaluated by probing logprobs for continuation tokens, the community quickly advanced to evaluating the open-ended completions of generative models directly. With the debut of OpenAI's O series of models, specialized thinking formats grew popular and specialized benchmarks such as AIME (static) and LiveBench and LiveCodeBench (dynamic) were developed to challenge thinking models.

\textbf{Agentic Models.} Most recently, the frontier has extended again, with benchmarks such as SWE-Bench and Terminal-Bench assigning scores to discrete tasks such as GitHub issue resolution~\citep{ merrill2026terminalbenchbenchmarkingagentshard, jimenez2024swebenchlanguagemodelsresolve}. Throughout this process, standardized evaluation platforms such as Evalchemy and Harbor have made benchmarks more affordable and reproducible~\citep{Harbor_Framework, evalchemy}. Over the past year and a half, as agentic AI has entered the research mainstream, various data curation strategies have been proposed for the effective post-training of AI agents, spanning both SFT and RL data curation~\citep{yang2025swesmithscalingdatasoftware, gandhi2026endlessterminalsscalingrl, pi2026dataengineeringscalingllm, shen2026serasoftverifiedefficientrepository, zeng2026davincidevagentnativemidtrainingsoftware, deepswe2025, cao2025skyrlagentefficientrltraining}. 
However, these works have two key limitations; firstly, they almost exclusively focus \textit{either} on SFT \textit{or} on RL, with little attention paid to how these steps intersect; secondly, they tend to focus on a single benchmark or a small cluster of closely related agentic benchmarks.
By contrast, our work focuses on generalization across agentic benchmarks, the interaction between SFT and RL, across several popular model scales, and utilizes a range of agentic benchmarks, including OpenThoughts-TB-Lite, new in this work. Recent works use Qwen-3.5 as base model for their experiments~\citep{ivison2026tmaxsimplerecipeterminal}. We have decided to keep Qwen-3 as base model to enable consistent comparisons throughout our project. Porting our data improvements to Qwen-3.5 and studying interactions between base model and SFT / RL data are important directions for future work.

\textbf{Public frameworks for building agents.} Although a detailed discussion is beyond the scope of this paper, we wish to emphasize that sound engineering practices constituted a central focus (and a central challenge) of this work, and acknowledge those we used. Many public frameworks for SFT are reasonably mature; we selected a fork of Llama-Factory, extending it to support ALST long-sequence training \citep{zheng-etal-2024-llamafactory,bekman2025arcticlongsequencetraining}. Our reinforcement learning framework was an extended version of the popular SkyRL framework; most of the improvements are described in \cite{novasky2026skyrlharbor}. We used \cite{Harbor_Framework} for environment, benchmark and harness management.

\section{SFT Data Pipeline}
\label{sec:sft-pipeline}

\providecommand{\subscriptval}[1]{{\color{gray!75!black}\scriptsize #1}}

\paragraph{Pipeline} This section introduces our experimental pipeline for building the OpenThoughts-Agent dataset. Our goal is to create the best dataset of (task, trajectory) pairs for supervised finetuning coding and terminal agents. The best dataset is the one that produces the highest-performing agent on downstream benchmarks. To do so, we ablate each pipeline step independently, and select the best-performing strategy based on the average z-score across three benchmarks. 
We compute the $z$-score of every candidate strategy's accuracy across the stage's full candidate set (subtracting the stage's per-benchmark mean and dividing by its standard deviation), then average the three resulting per-benchmark $z$-scores. This standardization gives each benchmark equal weight in the ranking despite differing accuracy ranges across the candidate set.
  Sections \ref{sec:task_generation_ablation} to \ref{sec:trajectory_filtering} describe each stage of the pipeline in detail.
 Unless otherwise specified, trajectories are generated by GLM-4.7-AWQ \citep{5team2025glm45agenticreasoningcoding} acting as the teacher in the \texttt{terminus-2} harness inside Daytona sandboxes. We conduct our experiments on datasets of size $10{,}000$ since that is small enough to be cost-effective yet large enough to provide a meaningful signal. 
\begin{figure}[t]
\centering
\includegraphics[width=0.8\textwidth]{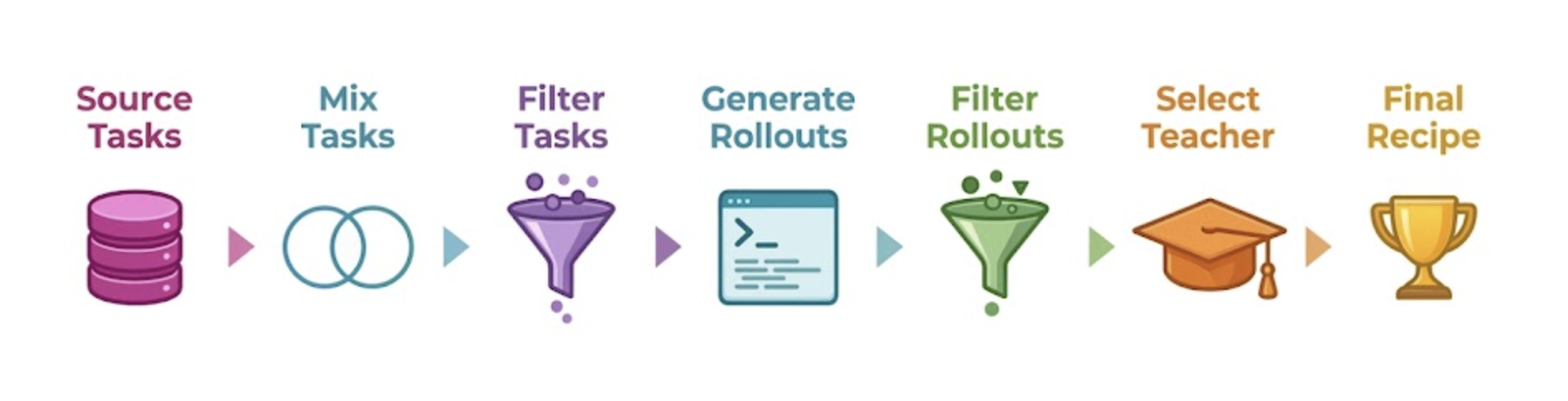}
\caption{\textbf{Six-stage SFT data pipeline for OpenThoughts-Agent.} Each stage is ablated independently in Sections~\ref{sec:task_generation_ablation}--\ref{sec:trajectory_filtering}. }
\label{fig:sft_pipeline}
\end{figure}

\paragraph{Training} For each ablation experiment, we utilize the full pipeline to generate $10{,}000$ trajectories for each data strategy, and we finetune Qwen3-8B \citep{qwen3technicalreport} on each dataset using full-parameter SFT, learning rate $4\mathrm{e}{-5}$ with cosine schedule, global batch size $96$, $7$ epochs, and   $32{,}768$ context length. Each $10{,}000$ finetune takes 160 GPU-hours on GH200s, allowing us to run dozens of pipeline ablations in parallel. These experiments inform the design choices for the final OpenThoughts-Agent pipeline. We include additional details on hyperparameters, training infrastructure, and per-stage SFT settings in the Appendix Section~\ref{app:sft-hparams}.

\paragraph{Evaluation Setup} 
We evaluate our models on a set of agent benchmarks that probe long-horizon software-engineering and terminal-use behavior. Our core suite consists of three benchmarks: (1) \textbf{OpenThoughts-TBLite} (100 tasks)~\citep{OpenThoughts-TBLite}: a curated collection of 100 Terminal-Bench style tasks that balanced across four difficulty buckets and engineered as a fast proxy for the full Terminal-Bench 2.0 performance, (2) \textbf{SWE-Bench Verified-100} (100 tasks)~\citep{jimenez2024swebenchlanguagemodelsresolve}: a stratified subsample (by repository) of the 500-task human-validated SWE-Bench Verified split; an agent produces a patch against a Python repo at a fixed commit, binary-scored by the upstream test harness, and (3) \textbf{Terminal Bench 2} (89 tasks)~\citep{merrill2026terminalbenchbenchmarkingagentshard}: hand-crafted, human-verified tasks that spans SWE, biology, security, system administration, and machine learning.

All evaluations run inside isolated Daytona sandboxes \citep{Daytona_Infrastructure_2025} using the \texttt{terminus-2} \citep{merrill2026terminalbenchbenchmarkingagentshard} agent harness, with $n=3$ stochastic re-runs per task and standard error reported across trials. To measure generalization, our pipeline experiments exclude a held-out set of out-of-distribution benchmarks, which are only measured once pipeline experimentation is complete. This held-out set consists of Aider Polyglot \citep{aiderpolyglot2024}, BFCL \citep{patil2025bfcl}, MedAgentBench \citep{jiang2025medagentbench}, GAIA \citep{mialon2023gaia}, and FinanceAgent-Terminal \citep{valsai2024financeagent}. \Cref{app:eval} contains further details on evaluation infrastructure and per-benchmark configuration.

\subsection{Sourcing Tasks}
\label{sec:task_generation_ablation}
\begin{table}[!htbp]
\centering
\resizebox{\textwidth}{!}{%
\begin{tabular}{rlcccccc}
\toprule
 & & \multicolumn{3}{c}{Benchmarks} & \multicolumn{2}{c}{Average} \\
\cmidrule(lr){3-5} \cmidrule(lr){6-7}
Rank & Strategy & SWE-bench Verified (100) & OT-TBLite & Terminal-Bench 2.0 & Raw & Normalized \\
\midrule
1 & SWESmith & \textbf{32.33}\,\subscriptval{1.83} & 17.63\,\subscriptval{1.63} & 6.37\,\subscriptval{1.18} & \textbf{18.78} & \textbf{+1.92} \\
2 & StackExchange SuperUser & 13.33\,\subscriptval{1.41} & 16.68\,\subscriptval{1.63} & \textbf{10.86}\,\subscriptval{1.35} & 13.62 & +1.51 \\
3 & StackExchange Tezos & 16.33\,\subscriptval{1.41} & 16.94\,\subscriptval{1.83} & 9.36\,\subscriptval{1.06} & 14.21 & +1.45 \\
4 & IssueTasks & 24.00\,\subscriptval{1.76} & 16.44\,\subscriptval{1.49} & 6.74\,\subscriptval{1.24} & 15.73 & +1.37 \\
\multicolumn{7}{c}{$\cdots$} \\
95 & AgentTuning-OS & 0.00\,\subscriptval{0.00} & 5.64\,\subscriptval{1.12} & 0.37\,\subscriptval{0.37} & 2.00 & -2.31 \\
\bottomrule
\end{tabular}}
\vspace{10pt}
\caption{\textbf{Task source choice has the largest spread of any pipeline stage.} Issue-resolution tasks and human-written infrastructure questions dominate the top, but improvements are spiky across benchmarks. Full ranking in Appendix~\ref{app:task-gen-full}.}
\label{tab:task_gen_table}
\end{table}
\FloatBarrier
Since an SFT agentic dataset comprises task descriptions and agent trajectory pairs, the strategy for generating task descriptions can change downstream accuracy by up to ~30 pp on SWE-Bench Verified-100 and ~10 pp on Terminal-Bench 2.0(\Cref{tab:task_gen_full_part1}, ranks 1 vs 95). The design space for task generation strategies is large, encompassing synthetic task generation, human task generation, and more. Moreover, the domain knowledge and skills covered by a task generation strategy can also lead to large differences in downstream performance.

To efficiently find the best task data generation strategies, we ablate a set of 95 task generation strategies covering different initial sources, modes of generation (synthetic vs.\ not synthetic), and knowledge domains for those tasks. We report the findings in \Cref{tab:task_gen_table}. The top-performing task generation strategies include synthetic issue-resolution tasks such as SWE-Smith and our own IssueTasks datasets, human-written computer-use questions such as StackExchange SuperUser and Tezos, and other strategies. The effect of choosing different task generation strategies leads to a large difference in downstream evals; for example, TerminalBench~2.0 scores range from 10.9\% to 0.4\%. We also observe that performance improvements at the top end of data generation strategies are spiky: the top coding-related datasets, such as SWE-Smith, greatly improve SWE-Bench, whereas the human-written infrastructure questions, such as StackExchange SuperUser, improve TerminalBench~2.0.

\subsection{Mixing Tasks}
\label{sec:mixing_tasks}

Our set of task descriptions will come from a mix of the task generation strategies from \Cref{sec:task_generation_ablation}. While there is a rich literature on data mixing strategies, we simply ablate the use of the top 1, top 2, and so on datasets in the mix, given the relative ranking of each task generation strategy from \Cref{sec:task_generation_ablation}. For the Top-$N$ mixing strategy, we take the top $N$ data generation strategies and sample $\frac{10{,}000}{N}$ task descriptions from each.

The results of mixing strategy are in \Cref{tab:mixing_strategies}. Mixing around the Top 4 or Top 8 task generation strategies works best and outperforms the non-mixed Top 1 baseline, since it performs well on all our benchmarks rather than solely on SWE-Bench.

\begin{table}[!htbp]
\centering
\resizebox{\textwidth}{!}{%
\begin{tabular}{rlcccccc}
\toprule
 & & \multicolumn{3}{c}{Benchmarks} & \multicolumn{2}{c}{Average} \\
\cmidrule(lr){3-5} \cmidrule(lr){6-7}
Rank & Mixing Strategy & SWE-Bench Verified (100) & OT-TBLite & Terminal-Bench 2.0 & Raw & Normalized \\
\midrule
1 & Top 4 & \textbf{29.33}\,\subscriptval{1.63} & \textbf{17.00}\,\subscriptval{1.71} & \textbf{8.24}\,\subscriptval{1.24} & \textbf{18.19} & \textbf{+0.49} \\
2 & Top 2 & \textbf{29.00}\,\subscriptval{1.60} & \textbf{18.12}\,\subscriptval{1.72} & \textbf{7.12}\,\subscriptval{1.06} & \textbf{18.08} & \textbf{+0.48} \\
3 & Top 8 & 28.00\,\subscriptval{1.70} & 15.86\,\subscriptval{1.70} & \textbf{8.61}\,\subscriptval{1.24} & \textbf{17.49} & \textbf{+0.19} \\
\multicolumn{7}{c}{$\cdots$} \\
6 & Top 1 & \textbf{30.67}\,\subscriptval{1.67} & 14.80\,\subscriptval{1.40} & 4.49\,\subscriptval{0.84} & 16.65 & -0.57 \\
\bottomrule
\end{tabular}}
\vspace{10pt}
\caption{\textbf{Mixing top-ranked task generation strategies, random shuffle within task.} Mixing the top-4 to top-8 strategies yields the strongest balanced performance, outperforming the unmixed top-1 baseline by avoiding over-specialization. }
\label{tab:mixing_strategies}
\end{table}
\FloatBarrier

\subsection{Task Augmentation Strategies}
\label{sec:task_augmentation}
\begin{table}[!t]
\centering
\resizebox{\textwidth}{!}{%
\begin{tabular}{rlcccccc}
\toprule
 & & \multicolumn{3}{c}{Benchmarks} & \multicolumn{2}{c}{Average} \\
\cmidrule(lr){3-5} \cmidrule(lr){6-7}
Rank & Augmentation Strategy & SWE-bench Verified (100) & OT-TBLite & Terminal-Bench 2.0 & Raw & Normalized \\
\midrule
1 & Original (no augmentation) & 20.67\,\subscriptval{1.56} & 17.22\,\subscriptval{1.48} & \textbf{8.99}\,\subscriptval{1.12} & \textbf{15.62} & \textbf{+0.39} \\
2 & Constrain $\rightarrow$ Harden & 17.33\,\subscriptval{1.49} & 17.56\,\subscriptval{1.59} & \textbf{9.36}\,\subscriptval{1.35} & \textbf{14.75} & \textbf{+0.27} \\
3 & Mixed (Original + Hardened) & 20.33\,\subscriptval{1.73} & 15.39\,\subscriptval{1.46} & \textbf{9.74}\,\subscriptval{1.40} & \textbf{15.15} & \textbf{+0.23} \\
4 & Mixed across sources & 15.33\,\subscriptval{1.33} & \textbf{19.02}\,\subscriptval{1.72} & \textbf{8.24}\,\subscriptval{1.24} & 14.20 & +0.10 \\
5 & Trace hints & \textbf{22.67}\,\subscriptval{1.63} & 17.17\,\subscriptval{1.48} & 5.99\,\subscriptval{1.12} & 15.27 & -0.13 \\
6 & Harden & 10.67\,\subscriptval{1.29} & 17.73\,\subscriptval{1.64} & \textbf{8.61}\,\subscriptval{1.18} & 12.34 & -0.41 \\
7 & Constrain & \textbf{24.33}\,\subscriptval{1.56} & 13.91\,\subscriptval{1.43} & 5.62\,\subscriptval{1.12} & 14.62 & -0.62 \\
\bottomrule
\end{tabular}}
\vspace{10pt}
\caption{\textbf{Task description augmentation strategies are within noise.} No LLM-driven augmentation strategy reliably outperforms the un-augmented baseline. }
\label{tab:task_augmentation}
\end{table}

After initially generating the task descriptions, a natural hypothesis is that refining them by clarifying their requirements or increasing their difficulty can improve dataset quality. We explore different methods for augmenting task descriptions, such as using an LLM to combine tasks from different sources, add new constraints to each task, harden the task descriptions, and more. The results of these interventions are in \Cref{tab:task_augmentation}. We find that all interventions fail to improve the baseline of leaving the task description untouched after generation.

\subsection{Filtering Tasks}
\label{sec:task_filtering}
\begin{table}[!t]
\centering
\resizebox{\textwidth}{!}{%
\begin{tabular}{rlcccccc}
\toprule
 & & \multicolumn{3}{c}{Benchmarks} & \multicolumn{2}{c}{Average} \\
\cmidrule(lr){3-5} \cmidrule(lr){6-7}
Rank & Filtering Strategy & SWE-bench Verified (100) & OT-TBLite & Terminal-Bench 2.0 & Raw & Normalized \\
\midrule
1 & Response Length (GPT-5, longest) & \textbf{22.67}\,\subscriptval{1.53} & \textbf{19.51}\,\subscriptval{1.72} & \textbf{10.11}\,\subscriptval{1.35} & \textbf{17.43} & \textbf{+1.38} \\
2 & Response Length (GPT-5, shortest) & \textbf{21.33}\,\subscriptval{1.67} & 16.91\,\subscriptval{1.65} & \textbf{9.74}\,\subscriptval{1.24} & 15.99 & +0.31 \\
3 & AskLLM & \textbf{20.67}\,\subscriptval{1.63} & 15.87\,\subscriptval{1.52} & \textbf{10.86}\,\subscriptval{1.18} & 15.80 & +0.19 \\
4 & Embedding diversity & 20.00\,\subscriptval{1.63} & 16.41\,\subscriptval{1.50} & 7.87\,\subscriptval{0.99} & 14.76 & -0.72 \\
5 & Random (baseline) & 19.67\,\subscriptval{1.70} & 14.77\,\subscriptval{1.36} & 7.87\,\subscriptval{1.35} & 14.10 & -1.17 \\
\bottomrule
\end{tabular}}
\vspace{10pt}
\caption{\textbf{Filtering task descriptions with LLM-based difficulty signals improves performance.} Selecting tasks for which GPT-5 produces longer responses improves over random selection by $\sim$3pp on average. }
\label{tab:task_filtering}
\end{table}
After determining the final mix of task generation strategies, filtering out poor tasks from a set of task descriptions can improve downstream performance. We repeat the ablation of the set of task description filters from OpenThoughts. \Cref{tab:task_filtering} holds the results. We find similar results: LLM task description filters yield the largest improvement among the filters we tested (+3 pp avg, \Cref{tab:task_filtering}). Filtering tasks to those that GPT-5 requires more tokens to solve finds tasks that lead to roughly 3 percentage points improvement across all benchmarks.

\subsection{Teacher Model}
\label{sec:teacher_ablation}
\begin{table}[!t]
\centering
\resizebox{\textwidth}{!}{%
\begin{tabular}{rlcccccc}
\toprule
 & & \multicolumn{3}{c}{Benchmarks} & \multicolumn{2}{c}{Average} \\
\cmidrule(lr){3-5} \cmidrule(lr){6-7}
Rank & Teacher Model & SWE-bench Verified (100) & OT-TBLite & Terminal-Bench 2.0 & Raw & Normalized \\
\midrule
1 & GLM 4.7 (Quantized) & 28.00\,\subscriptval{1.76} & \textbf{17.86}\,\subscriptval{1.60} & \textbf{8.61}\,\subscriptval{1.40} & \textbf{18.16} & \textbf{+0.73} \\
2 & Kimi K2.5 & \textbf{33.33}\,\subscriptval{1.83} & 14.19\,\subscriptval{1.51} & \textbf{8.24}\,\subscriptval{1.06} & \textbf{18.59} & \textbf{+0.66} \\
3 & GLM 5 & \textbf{33.00}\,\subscriptval{1.60} & 14.30\,\subscriptval{1.78} & \textbf{7.50}\,\subscriptval{1.18} & \textbf{18.27} & \textbf{+0.66} \\
4 & GLM 4.6 (Quantized) & 26.00\,\subscriptval{1.67} & \textbf{16.99}\,\subscriptval{1.80} & 6.37\,\subscriptval{1.24} & 16.45 & +0.08 \\
5 & GPT-5.3-Codex & 21.67\,\subscriptval{1.33} & 10.42\,\subscriptval{1.55} & 3.75\,\subscriptval{0.99} & 11.94 & -1.47 \\
\bottomrule
\end{tabular}}
\vspace{10pt}
\caption{\textbf{Teacher model ablation: stronger model $\neq$ better teacher.} Despite GPT-5.3-Codex being the strongest model on these benchmarks, it is the weakest teacher, underperforming GLM 4.7 AWQ by $\sim$5pp on Terminal-Bench~2.0. }
\label{tab:teacher_ablation}
\end{table}

After selecting the set of methods for generating task descriptions, we ablate the design choices for generating the agentic rollouts. Prior work has shown that the choice of teacher can make a significant difference in downstream evals~\citep{guha2025openthoughtsdatarecipesreasoning}. Building on this awareness, we take the best mix of task descriptions from \Cref{sec:mixing_tasks} and generate a dataset with different teachers that perform well on TerminalBench~2.0, including GPT-5.3-Codex, Kimi K2.5, GLM-4.6-AWQ, GLM 5, and our baseline GLM-4.7-AWQ. Our results are in \Cref{tab:teacher_ablation}. Despite GPT-5.3-Codex being the best-performing model, it is a worse teacher than GLM-4.7-AWQ, representing a roughly 5\% decrease in performance on TerminalBench~2.0. The best teacher is GLM 4.7, despite being older and less performant (e.g., on Terminal Bench) than Kimi K2.5.

\subsection{Filtering Agent Rollouts}
\label{sec:trajectory_filtering}
\begin{table}[!t]
\centering
\resizebox{\textwidth}{!}{%
\begin{tabular}{rlcccccc}
\toprule
 & & \multicolumn{3}{c}{Benchmarks} & \multicolumn{2}{c}{Average} \\
\cmidrule(lr){3-5} \cmidrule(lr){6-7}
Rank & Trajectory Filter & SWE-bench Verified (100) & OT-TBLite & Terminal-Bench 2.0 & Raw & Normalized \\
\midrule
1 & Min turns $\geq 5$ & \textbf{29.00}\,\subscriptval{1.56} & \textbf{19.10}\,\subscriptval{1.57} & \textbf{11.61}\,\subscriptval{1.30} & \textbf{19.90} & \textbf{+1.25} \\
2 & Filter timeouts & 26.67\,\subscriptval{1.70} & \textbf{18.03}\,\subscriptval{1.60} & \textbf{10.49}\,\subscriptval{1.18} & 18.39 & -0.35 \\
3 & Filter subagent traces & 23.00\,\subscriptval{1.63} & \textbf{17.31}\,\subscriptval{1.58} & \textbf{10.86}\,\subscriptval{1.35} & 17.06 & -0.90 \\
\bottomrule
\end{tabular}}
\vspace{10pt}
\caption{\textbf{Filtering agent rollouts: keeping longer trajectories helps.} The minimum-turns filter outperforms timeout and subagent filters across all three benchmarks.}
\label{tab:answer_filtering}
\end{table}

Filtering out lower-quality agentic rollouts is one way to improve dataset quality. We apply simple heuristic filters, including removing traces that hit a timeout during generation, removing subagent traces, and removing traces with fewer than 5 turns. Our results are in \Cref{tab:answer_filtering}. We again find that methods that yield longer agentic traces yield better performance, with filtering traces with fewer than 5 turns yielding the largest improvement in downstream evaluation scores.
Because longer traces also carry more tokens, we verify in \Cref{app:compute-controlled-filtering} that this gain persists at a matched token budget, confirming it stems from higher-quality multi-turn supervision rather than additional training compute.
\section{Scaling Up SFT Data}
\label{sec:scaling}

Scaling dataset size is a powerful method for improving downstream performance. After the pipeline ablations, our final pipeline yields a dataset of 10K datapoints. To increase the dataset size, several simple strategies are possible: (1) using the same task descriptions and generating more agentic rollouts per task; (2) using more task descriptions from the original sources; (3) using synthetic augmentation to create new task descriptions; and (4) using more initial sources.

\begin{figure}[!t]
\centering
\includegraphics[width=\textwidth]{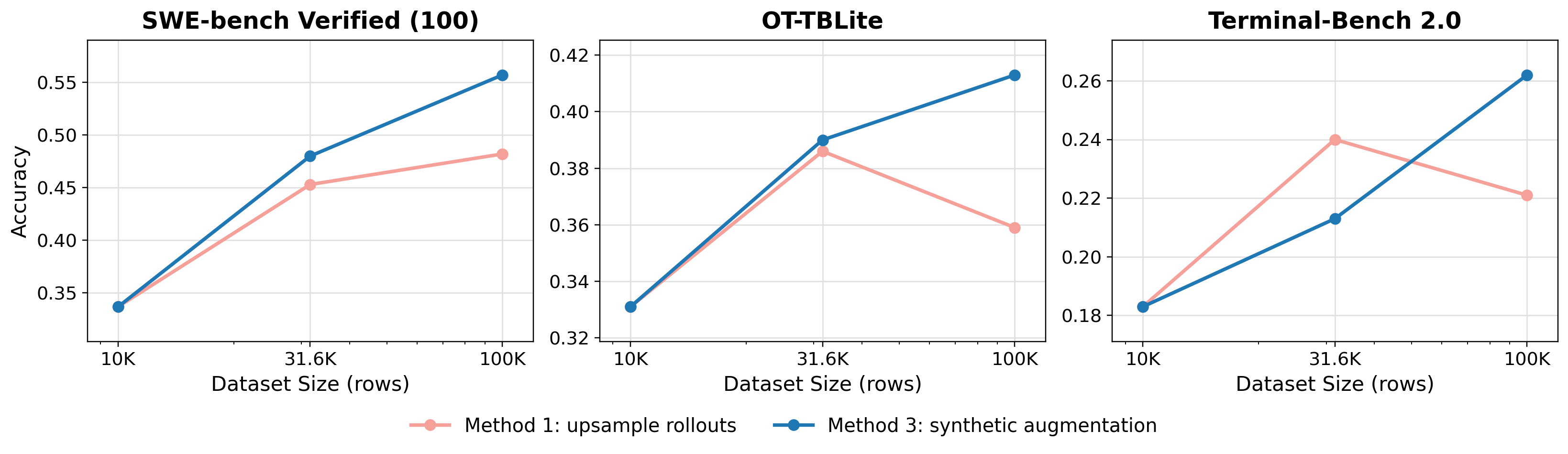}
\caption{\textbf{Synthetic augmentation scales past the upsampling plateau.} Both methods build on the same 10K base and diverge only when scaling beyond it. Method~1 (upsampling additional rollouts per task description) plateaus from 31.6K to 100K, while Method~3 (synthetic task augmentation) continues to improve on all three benchmarks. Error bars are standard error across three stochastic re-runs.}
\label{fig:scaling_methods}
\end{figure}

We start with Method~1 since it is the simplest. The results are in \Cref{fig:scaling_methods}. We find that performance plateaus from 31.6K to 100K (+3pp on SWE-Bench Verified-100, $-2$pp on Terminal-Bench~2.0; both within standard error), suggesting that task-description diversity is the bottleneck. While Method~2 would be a simple fix, we are limited by our initial sources; for example, Tezos contains only 997 unique task descriptions.

\FloatBarrier

We attempt sourcing from more tasks from Top-4, Top-8, and Top-16  datasets at the 100K scale  (Method~4 results in \Cref{tab:top16_negative}). Adding more sources beyond Top-4 does \emph{not} reliably help: Top-8 does not significantly outperform Top-4 on every benchmark, while broadening to Top-16 \emph{hurts} on every benchmark. We therefore retain the original Top-4 source mix for the remainder of our experiments and do not pursue Method~4 further.

\begin{table}[!t]
\centering
\small
\begin{tabular}{lccc}
\toprule
Source Mix & SWE-Bench Verified-100 & OT-TBLite & Terminal-Bench~2.0 \\
\midrule
Top-4  & 45.33\,\subscriptval{1.73} & 36.90\,\subscriptval{1.82} & 21.72\,\subscriptval{1.54} \\
Top-8  & \textbf{49.00}\,\subscriptval{1.45} & \textbf{38.87}\,\subscriptval{2.09} & \textbf{22.85}\,\subscriptval{1.30} \\
Top-16 & 40.33\,\subscriptval{1.41} & 33.14\,\subscriptval{2.01} & 20.60\,\subscriptval{1.67} \\
\bottomrule
\end{tabular}
\vspace{6pt}
\caption{ \textbf{Task-Source diversity has a negligible effect at larger data scales.} Adding sources beyond Top-4 does not reliably lift performance. All rows are 32B SFT models trained on a 100K-row dataset with the $\geq 5$ turns trace filter.}
\label{tab:top16_negative}
\end{table}
\FloatBarrier

We turn to synthetic task augmentation (Method 3) and report the results in \Cref{fig:scaling_methods}. Concretely, we take the four highest-scoring sources from \Cref{sec:mixing_tasks} (swe-smith, stackexchange-superuser, stackexchange-tezos, and issue-tasks) and replace the Tezos subset with a synthetically augmented version of itself. Tezos contributes the fewest unique task descriptions (only 997 unique tasks). We replace it with a synthetically augmented version of itself: we apply the instruction-rewriting strategies from  \Cref{sec:task_augmentation} to those same 997 base problems, expanding their distinct surface forms from $\sim$902 to over 21K without introducing any new underlying problems. We then use the \texttt{gpt-5-nano} response-length signal from \Cref{sec:task_filtering} as upsampling \emph{weights} rather than as a hard top-$k$ filter — every unique task receives at least one rollout, with the remaining capacity allocated proportionally to score. Our goal is to preserve full task coverage at every dataset scale. We then apply the $\geq 5$-turn trace filter uniformly across all four sources. We see continued performance gains at larger dataset scales, demonstrating that augmentation overcomes the limited task-description-diversity bottleneck observed in Method~1.

\FloatBarrier

\paragraph{OpenThoughts-Agent-v2}

\begin{figure}[!t]
\centering
\includegraphics[width=1.0\textwidth]{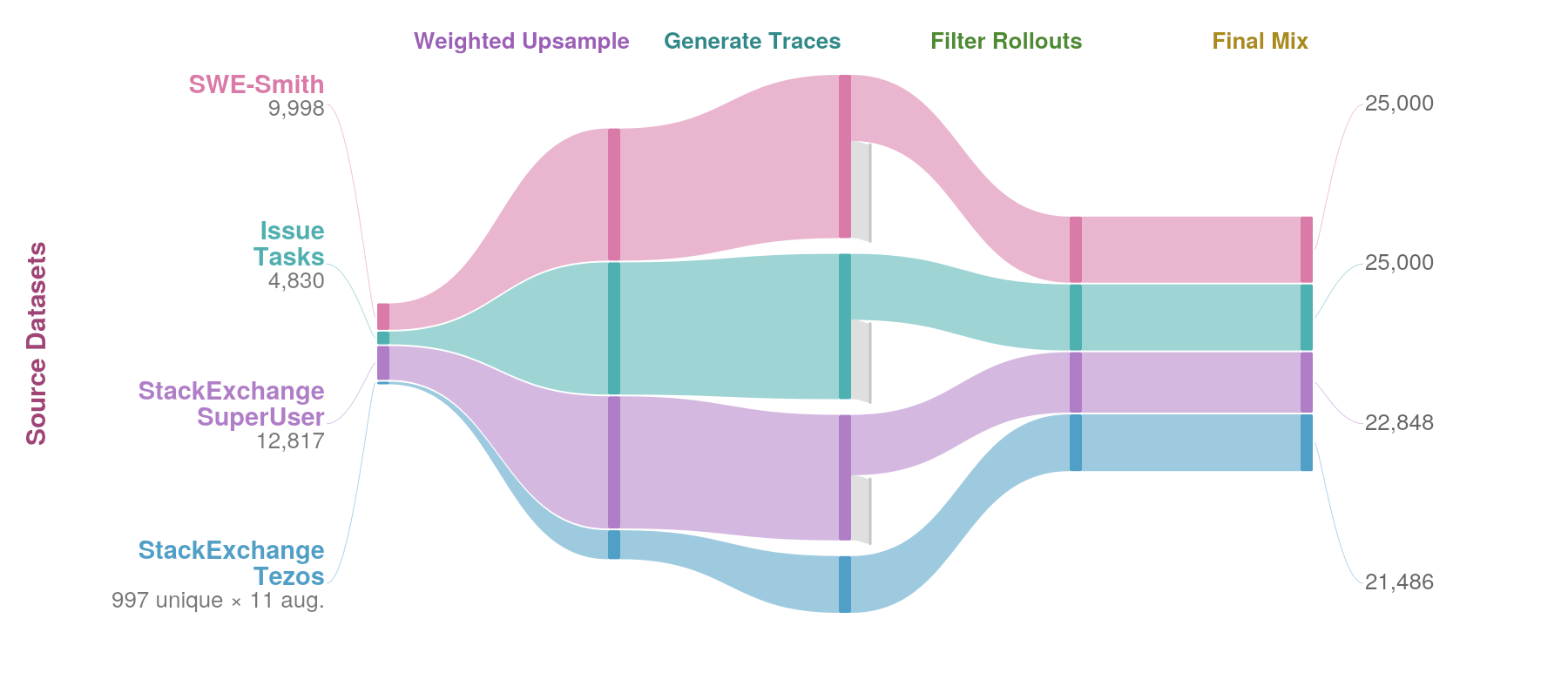}
\caption{\textbf{OpenThoughts-Agent Full Data Pipeline.} Our final SFT dataset is 100k agentic traces.}
\label{fig:sankey}
\end{figure}

At the 100K scale, we observe the best performance for the 32B SFT model, reaching \textbf{26.2\%} on Terminal-Bench~2.0, \textbf{41.3\%} on OT-TBLite, and \textbf{55.7\%} on SWE-Bench Verified-100, showing a monotonic improvement from 31.6K of $+7.7$pp on SWE-Bench Verified-100 and $+5.0$pp on Terminal-Bench~2.0. We report the full data generation pipeline for our final 100k-sized dataset, \textbf{OpenThoughts-Agent-v2} in \Cref{fig:sankey}. We begin from our initial 4 task sources, including synthetic GitHub Issues, human-written Linux tasks, and human-written cryptocurrency questions. We repeat task descriptions and synthetically augment the Tezos questions. We generate agentic rollouts using GLM-4.7-AWQ and filter out traces with fewer than 5 turns.

\section{Reinforcement Learning}
\label{sec:rl-data}

Many high-profile RL datasets, including SWE-Smith and R2EGym~\citep{yang2025swesmithscalingdatasoftware,jain2025r2egymproceduralenvironmentshybrid}, followed a similar approach: select a representative GitHub repository whose state and dependencies can be captured in a Docker container, select or synthesize a flawed commit with failing tests, and generate a natural-language problem statement describing the issue and requesting the agent turn the failing tests green. We incorporate these efforts into a larger-scale and more systematic series of ablation studies -- similar to our work in the SFT domain, we investigate to what extent model performance depends on the \emph{source} of the RL training data. To isolate this, we run a controlled data-source ablation, which we describe below.

\subsection{Experimental Details}

To control compute consumption, we focus our RL investigations in the 8B regime. We run async RL using the RLOO algorithm \citep{ahmadian2024back} with standard binary rewards on verifier success (expect PASS : PASS, expect FAIL : FAIL for all tests). We conduct RL starting from a distilled 8B checkpoint (OT-Agent-ColdSFT) trained on SWE-Smith traces generated by a GLM 4.7 AWQ teacher with thinking, which we also train. Our hero run trained on 24xA100 80GB GPUs with a batch size of 64 and a total wallclock time of $\approx$46\,hours. For additional technical details and complete hyperparameters, please refer to \Cref{app:rl-hparams}.

\subsection{Sourcing Tasks in RL}
\label{sec:rl-data-ablation}

\begin{table}[!t]
\centering
\resizebox{\textwidth}{!}{%
\begin{tabular}{rlcccccc}
\toprule
 & & \multicolumn{3}{c}{Benchmarks} & \multicolumn{2}{c}{Average} \\
\cmidrule(lr){3-5} \cmidrule(lr){6-7}
Rank & RL Data Source & SWE-bench Verified (100) & OT-TBLite & Terminal-Bench 2.0 & Raw & Normalized \\
\midrule
1 & \texttt{pymethods2test} & \textbf{35.67}\,\subscriptval{1.83} & \textbf{16.02}\,\subscriptval{1.58} & \textbf{13.48}\,\subscriptval{1.50} & \textbf{21.72} & \textbf{+1.73} \\
2 & \texttt{r2egym} & 28.67\,\subscriptval{1.67} & \textbf{16.84}\,\subscriptval{1.64} & 6.74\,\subscriptval{1.45} & \textbf{17.42} & \textbf{+0.50} \\
3 & \texttt{nemotron-code-oracle} & 25.00\,\subscriptval{1.86} & \textbf{16.78}\,\subscriptval{1.67} & 6.74\,\subscriptval{1.24} & \textbf{16.17} & \textbf{+0.22} \\
4 & \texttt{llm-verifier-freelancer} & 22.33\,\subscriptval{1.53} & \textbf{14.87}\,\subscriptval{1.27} & 8.61\,\subscriptval{1.35} & 15.27 & -0.24 \\
5 & \texttt{inferredbugs} & 26.00\,\subscriptval{1.67} & 14.30\,\subscriptval{1.55} & 6.37\,\subscriptval{0.99} & 15.56 & -0.45 \\
6 & \texttt{swesmith} & 24.33\,\subscriptval{1.73} & 14.30\,\subscriptval{1.51} & 6.74\,\subscriptval{1.12} & 15.12 & -0.51 \\
7 & \texttt{code-contests} & 23.67\,\subscriptval{1.67} & 13.75\,\subscriptval{1.55} & 7.87\,\subscriptval{1.12} & 15.10 & -0.56 \\
8 & \texttt{nl2bash} & 21.00\,\subscriptval{1.49} & 14.51\,\subscriptval{1.44} & 6.74\,\subscriptval{1.12} & 14.08 & -0.70 \\
\bottomrule
\end{tabular}}
\vspace{10pt}
\caption{\textbf{Data source strongly influences agentic RL performance.} Across eight 8B RL runs that hold the training pipeline fixed and vary only the source, performance varies well beyond noise.}
\label{tab:rl_data_ablation}
\end{table}

In our ablation, every run uses identical hyperparameters and evaluation criteria, the only thing we vary is the dataset the agent trains on. We compare six sources spanning competitive programming recast as Python contracts (\texttt{pymethods2test}), real-repository bug-fixing (\texttt{inferredbugs}), competitive-programming environments (\texttt{code-contests}), an LLM-filtered Nemotron code-oracle mix (\texttt{nemotron-code-oracle}), an LLM-verified freelancer-task set (\texttt{llm-verifier-freelancer}), and natural-language-to-Bash tasks (\texttt{nl2bash}), as well as SWE-Smith and R2EGym, with the same evaluation conditions described in \Cref{sec:sft-pipeline}. The training logs indicate all sources are learnable to some degree -- \texttt{inferredbugs} and \texttt{nemotron-code-oracle} runs show healthy, monotonically rising mean reward over training (roughly $0.21\!\to\!0.46$ and $0.06\!\to\!0.36$ across their exported steps, respectively), whereas \texttt{code-contests} plateaus at a much lower reward ceiling ($0.06\!\to\!0.14$)

\textbf{The data source matters.} \Cref{tab:rl_data_ablation} shows that source ablation spans a $7.6$-point range in raw average accuracy, larger than the $2.0$-point run-to-run reproducibility variance (see \Cref{tab:rl_reproducibility}), but smaller than the variance derived from SFT source ablation as described in \Cref{sec:sft-pipeline}.

Our strongest source, \texttt{pymethods2test}, is a mix of Codeforces / CodeChef / TopCoder style competitive-programming problems that have been re-cast as single-function Python contracts with synthesized docstring-style task descriptions and auto-generated unittest suites. There is no multi-file editing, no repository navigation, no shell-state accumulation across turns. Skills exercised include 1D/2D dynamic
programming, string and pattern matching (KMP), combinatorics, matrix/grid construction, ad-hoc puzzles, and formatted table/string generation. Reference solutions average around 20 LOC and task description around 200 words. This dataset is highly reproducible (no potentially stale github references), highly usable (all tasks use the same build environment) and has a clear, and in this case appropriately moderate, difficulty ceiling. The concise but challenging source tasks induce the cold-start model to adopt a consistent problem-solving pattern; during RL, it replaces loops of thinking and exploratory (sed and grep) tool calls with a compact explore, patch, and submit policy. We further analyze the emergent behavioral changes behind the \texttt{pymethods2test} result -- and why its RL signal pushes the policy to explore more aggressively than the alternatives -- in \Cref{app:rl-behavior}. 

Among the alternatives, the synthetic-and-competitive-programming sources (\texttt{inferredbugs}, \texttt{code-contests}) lead on ID, while the more heterogeneous tool-use sources (\texttt{llm-verifier-freelancer}, \texttt{nl2bash}) are competitive on OOD despite weaker ID scores. The moderate ID / OOD decoupling suggests that data sources emphasizing single-function code correctness transfer most cleanly to the SWE/terminal Core benchmarks, while broader tool-use data buys OOD generalization at some ID cost; only \texttt{pymethods2test} sits at the top of both.

\subsection{Results}
\label{sec:rl-results}

\begin{table}[t!]
\centering
\newcolumntype{C}[1]{>{\centering\arraybackslash}p{#1}}
\resizebox{\textwidth}{!}{
\begin{tabular}{p{0.15cm} l | C{1.4cm}C{1.4cm}C{1.4cm}C{1.4cm}C{1.4cm}C{1.4cm}
@{\hspace{0.3cm}}p{0.1cm}@{\hspace{0cm}} C{1.4cm}}
\toprule  
& Benchmark & \rotatebox{45}{\textbf{OT-Agent-ColdSFT+RL-8B}} &
\rotatebox{45}{\textbf{OT-Agent-SFT-8B (100K)}} & \rotatebox{45}{\textbf{Nemotron-Terminal-8B}} &
\rotatebox{45}{\textbf{OT-Agent-SFT-8B (10K)}} & \rotatebox{45}{\textbf{SWE-Lego-Qwen3-8B}} &
\rotatebox{45}{\textbf{Endless Terminals}} & \rotatebox{45}{\scriptsize{------ Base Model ------}} &
\rotatebox{45}{\textbf{Qwen3-8B}} \\
\midrule
& Base Model & {\raisebox{-.5\height}{\includegraphics[scale=.1]{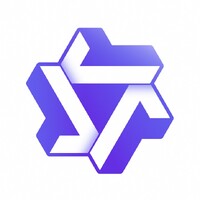}}} &
{\raisebox{-.5\height}{\includegraphics[scale=.1]{qwen_logo.jpeg}}} &
{\raisebox{-.5\height}{\includegraphics[scale=.1]{qwen_logo.jpeg}}} &
{\raisebox{-.5\height}{\includegraphics[scale=.1]{qwen_logo.jpeg}}} &
{\raisebox{-.5\height}{\includegraphics[scale=.1]{qwen_logo.jpeg}}} &
{\raisebox{-.5\height}{\includegraphics[scale=.1]{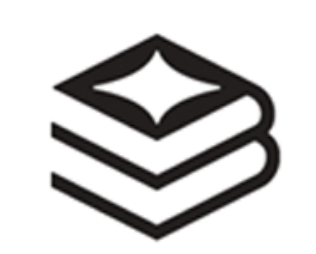}}} & \multicolumn{1}{!{\vrule width 0.5pt}c}{} &
{N/A} \\
& Train Size & {10K+5K} & {100K} & {264K} & {10K} & {18K} & {15K+3K} & \multicolumn{1}{!{\vrule width 0.5pt}c}{} &
{N/A} \\
& Method & {SFT+RL} & {SFT} & {SFT} & {SFT} & {SFT} & {SFT+RL} & \multicolumn{1}{!{\vrule width 0.5pt}c}{} &
{N/A} \\
\midrule  
\multirow{1}{*}{} & Average & \textbf{27.9} & 27.4 & 26.0 & 24.3 & 18.5 & 17.4 &
\multicolumn{1}{!{\vrule width 0.5pt}c}{} & 10.2 \\
\midrule
\multirow{2}{*}{\rotatebox{90}{\textit{Core}}} & SWE-bench Verified & 31.9 & \textbf{38.9} & 22.1 & 22.7 & 37.6
& 12.5 & \multicolumn{1}{!{\vrule width 0.5pt}c}{} & 13.2 \\
& Terminal-Bench 2.0 & \textbf{13.5} & 10.9 & 13.1 & 7.9 & 0.7 & 8.2 & \multicolumn{1}{!{\vrule width
0.5pt}c}{} & 2.2 \\
\midrule
\multirow{5}{*}{\rotatebox{90}{\textit{OOD}}} & Aider Polyglot & 18.1 & 15.9 & 12.6 & 14.2 & 3.7 &
\textbf{18.8} & \multicolumn{1}{!{\vrule width 0.5pt}c}{} & 11.7 \\
& BFCL-Parity & 71.5 & 65.9 & 59.9 & \textbf{79.4} & 75.3 & 48.8 & \multicolumn{1}{!{\vrule width 0.5pt}c}{} &
34.4 \\
& MedAgentBench & 31.1 & 36.2 & \textbf{48.6} & 25.8 & 5.0 & 18.4 & \multicolumn{1}{!{\vrule width 0.5pt}c}{}
& 3.8 \\
& GAIA-127 & 6.8 & 6.6 & \textbf{14.4} & 5.5 & 6.0 & 5.2 & \multicolumn{1}{!{\vrule width 0.5pt}c}{} &
5.0 \\
& FinanceAgent-Terminal & \textbf{22.7} & 17.3 & 11.3 & 14.7 & 1.3 & 10.0 & \multicolumn{1}{!{\vrule width
0.5pt}c}{} & 1.3 \\
\bottomrule
\end{tabular}}
\vspace{6pt}
\caption{\textbf{RL main results (8B scale).} Each cell reports the highest accuracy (\%) attained by the model across
all evaluated agent harnesses for that benchmark. \textbf{Average} is the mean of evaluated cells per column over the
seven benchmarks reported here. \raisebox{-.03cm}{\includegraphics[scale=.055]{qwen_logo.jpeg}} denotes a model
trained from Qwen3-8B. \raisebox{-.03cm}{\includegraphics[scale=.055]{ot_logo.png}} denotes a model trained
from OpenThinker-Agent-v1-SFT. }
\label{tab:ot_agent_8B_table3}
\end{table}

In \Cref{tab:ot_agent_8B_table3}, we show that the complete post-training pipeline (SFT + RL) outperforms the baslines on both core benchmarks, as well as on average across the entire benchmark suite, improving by 18 points, on average, above the Qwen3-8B base model. 

\begin{table}[!htbp]
\centering
\resizebox{\textwidth}{!}{%
\begin{tabular}{rlcccccc}
\toprule
& & \multicolumn{3}{c}{Benchmarks} & \multicolumn{2}{c}{Average} \\
\cmidrule(lr){3-5} \cmidrule(lr){6-7}
Rank & Model (training recipe) & SWE-bench Verified (100) & OT-TBLite & Terminal-Bench 2.0 & Raw & Normalized \\
\midrule
1 & OT-Agent-ColdSFT+RL-8B & \textbf{35.7}\,\subscriptval{1.8} & \textbf{16.0}\,\subscriptval{1.6} &
\textbf{13.5}\,\subscriptval{1.5} & \textbf{21.7} & \textbf{+1.2} \\

2 & OT-Agent-SFT-8B (10K) & 24.3\,\subscriptval{1.9} & 15.6\,\subscriptval{1.5} & 7.9\,\subscriptval{1.4} & 15.9 & +0.5 \\

3 & OT-Agent-ColdSFT & 23.7\,\subscriptval{1.6} & 14.8\,\subscriptval{1.4} & 6.7\,\subscriptval{1.1} & 15.1 & +0.4 \\
4 & RL on Qwen3-8B (no SFT) & 1.0\,\subscriptval{0.5} & 7.8\,\subscriptval{1.1} & 1.9\,\subscriptval{0.7} & 3.6 & -0.9
\\
5 & Qwen3-8B & 5.3\,\subscriptval{1.1} & 1.3\,\subscriptval{0.3} & 1.5\,\subscriptval{0.4} & 2.7 & -1.1 \\
\bottomrule
\end{tabular}}
\vspace{10pt}
\caption{\textbf{RL on top of moderately strong SFT outperforms other strategies.} Our SFT + RL model (1), which starts
from a medium-strength SFT base model (3), outperforms both RL from a weaker SFT base model (4) and SFT alone (3).
Each cell reports the highest accuracy (\%) attained in the Terminus-2 harness. Normalized = mean of
per-benchmark z-scores across 5 entries.}
\label{tab:8b_pipeline_ablation}
\end{table}

\textbf{``Undertrained'' SFT models benefit more from RL, and ultimately outperform pure-distilled models and RL-only models.} Consistent with prior work, we find in \Cref{tab:8b_pipeline_ablation} that RL provides the most gains when the SFT model is selected with RL in mind~\citep{kang2025quagmiressftrlposttraininghigh}. The Qwen3-8B model, which performs poorly in the \texttt{terminus-2} harness on agentic benchmarks, is unable to benefit from agentic RL. The model SFT'd on a smaller amount of data provides the better starting point.

\section{Conclusion}
\label{sec:conclusion}

Agents are increasingly central to science and technology, yet little is publicly known about the data curation techniques used to train them.
We address this gap by conducting controlled ablations on a six-stage SFT data curation pipeline, alongside a focused study of agentic RL data. Using the resulting dataset, OpenThoughts-Agent-v2, we finetune Qwen3-32B into OpenThinker-Agent-32B, the strongest open-data <=32B model (Qwen3 family or earlier) on the average of seven agentic benchmarks spanning software engineering, terminal use, tool calling, healthcare, finance, and general assistant tasks. At the 8B scale, combining our SFT data with our \texttt{pymethods2test} RL dataset further outperforms the strongest existing <=8B baselines, providing initial evidence that the SFT and RL stages of agentic post-training can be designed to compose. 
We release the data, pipeline, and models at \url{https://openthoughts.ai} to enable broader open research on agentic models.

Limitations include our RL investigation which was conducted only at the 8B scale due to compute constraints; whether the same RL recipe transfers to the 32B regime remains an open question.  We also do not ablate the choice of base model: all SFT runs begin from the Qwen3 family, so the contribution of base-model pretraining to the final performance is not isolated. Finally, our largest training set contains 100K trajectories; whether the trends we observe extrapolate to multi-million-trajectory regimes remains untested.

\paragraph{Broader Impacts}
This work aims to advance open research on agentic models by releasing the dataset, pipeline, and trained models. Open release accelerates scientific progress and lowers the barrier to entry for academic and independent research. However, agentic models are inherently dual-use technologies: the same capabilities that enable beneficial automation can also enable misuse, including unauthorized actions on shared systems. We encourage downstream users to deploy these models with appropriate sandboxing and human oversight.

\begin{ack}
\label{sec:acknowledgements}

MN and JJ acknowledge funding by EU Horizon under grant no. 101214398 (ELLIOT) and co-funding by EU from Digital Europe Programme under grant no. 101195233 (openEuroLLM), co-funding from EU under Digital Europe Programme under grant no. 101198470 (LLMs4EU) and from EuroHPC Joint Undertaking programme under grant no. 101182737 (MINERVA), funding by the German Federal Ministry of Research, Technology and Space (BMFTR) under grant no. 01IS24085C (OPENHAFM), under the grant 01IS22094B (WestAI -- AI Service Center West), and under the grant 16HPC117K (MINERVA).

BF gratefully acknolwedges resources of the Oak Ridge Leadership Computing Facility (OLCF) and Argonne Leadership Computing Facility (ALCF)] which are a DOE Office of Science User Facility. This work was supported by an award from the ASCR Leadership Computing Challenge (ALCC) under project ERCAP0034861, and the ongoing support of Oumi.AI.

LS gratefully acknowledges the Open Philanthropy Institute for Foundations of Machine Learning (IFML), Apple, and the Microsoft Grant in Customer Experience Innovation for their support on this project.

The entire team wishes to acknowledge the Gauss Centre for Supercomputing e.V. (GCS) for funding this work by providing computing time through the John von Neumann Institute for Computing (NIC) on the supercomputer JUWELS Booster and JUPITER at J\"ulich Supercomputing Centre (JSC), EuroHPC Joint Undertaking for computing time and storage on the EuroHPC supercomputer LEONARDO, hosted by CINECA (Bologna, Italy) and the LEONARDO consortium through an EuroHPC AI Factory Science and Innovation grant EHPC-AIF-2025SC04-290 and on EuroHPC supercomputer MareNostrum5 hosted by BSC (Barcelona, Spain) through EuroHPC development access grant EHPC-DEV-2026D01-097, storage resources on JUST granted and operated by JSC and supported by Helmholtz Data Federation (HDF), computing time granted by the JARA and JSC on the supercomputer JURECA at JSC, computing time granted on prototype JEDI via JUREAP (JUPITER Early Access Program) grant at JSC and computing time granted via Gauss AI Competition (reformo) on JUPITER through GCS and German Federal Ministry of Research, Technology and Space (BMFTR). LAION further acknowledges public storage grant by HuggingFace that allows us to provide convenient access to the output of the open-source research to broad community via HF repository. Further thanks go for support provided by supercomputing facilities and their teams, especially to Bjoern Hagemeier and Mathis Bode from Juelich Supercomputer Center (JSC, Germany). This project also benefited from the support of the TACC, NYU Torch, and ZIH Capella supercompute clusters, as well as Modal and Google, and Anyscale for hosting our group meetings. 

Finally, we owe a deep debt of gratitude to Daytona.io for providing a robust and scalable container solution for our agentic post-training experiments, and to the Laude Institute for supporting our project with a Slingshots // TWO grant, and the Harbor Framework for allowing us to provide input in the development of the their sandboxing evaluation and optimization environment.
\end{ack}

\bibliography{otagent}
\bibliographystyle{plain}


\appendix

\newpage
\clearpage

\startcontents[sections]
\printcontents[sections]{l}{1}{\setcounter{tocdepth}{3}}

\clearpage

\section{Full SFT Pipeline Ablation Tables}
\label{app:sft-ablations-full}

This appendix contains the full ablation tables for the SFT pipeline experiments summarized in \Cref{sec:sft-pipeline}. All experiments follow the setup described in \Cref{sec:sft-pipeline}: $10{,}000$ trajectories per dataset, Qwen3-8B fine-tuned with full-parameter SFT, and three stochastic re-runs per benchmark with standard error reported as a subscript.

\subsection{Task Generation Strategies (Full Ranking)}
\label{app:task-gen-full}

\Cref{tab:task_gen_full_part1} and \Cref{tab:task_gen_full_part2} report the full ranking of all 95 task generation strategies summarized in \Cref{tab:task_gen_table}. Strategies are sorted by normalized average $z$-score across the three benchmarks.

\begin{table}[!htbp]
\centering
\footnotesize
\setlength{\tabcolsep}{4pt}
\begin{tabular}{rlccccc}
\toprule
 & & \multicolumn{3}{c}{Benchmarks (\%)} & \multicolumn{2}{c}{Average} \\
\cmidrule(lr){3-5} \cmidrule(lr){6-7}
Rank & Strategy & SWE-Bench Verified (100) & OT-TBLite & Terminal-Bench 2.0 & Raw & Normalized \\
\midrule
1  & swe-smith                       & 32.33 & 17.63 &  6.37 & 18.78 & +1.92 \\
2  & stackexchange-superuser        & 13.33 & 16.68 & 10.86 & 13.62 & +1.51 \\
3  & stackexchange-tezos            & 16.33 & 16.94 &  9.36 & 14.21 & +1.45 \\
4  & issue-tasks                    & 24.00 & 16.44 &  6.74 & 15.73 & +1.37 \\
5  & repo-scaffold                  & 11.00 & 20.68 &  8.24 & 13.31 & +1.31 \\
6  & r2egym                         & 28.33 & 16.57 &  4.12 & 16.34 & +1.17 \\
7  & stackexchange-tor              & 15.33 & 16.11 &  8.61 & 13.35 & +1.16 \\
8  & swegym                         & 27.00 & 17.09 &  4.12 & 16.07 & +1.14 \\
9  & code-feedback                  & 11.00 & 17.39 &  8.61 & 12.33 & +1.04 \\
10 & stackexchange-unix             & 16.00 & 18.90 &  5.99 & 13.63 & +1.02 \\
11 & taskmaster2                    & 10.00 & 16.44 &  8.99 & 11.81 & +0.95 \\
12 & staqc                          & 11.00 & 19.41 &  6.74 & 12.38 & +0.91 \\
13 & multifile-composition          &  9.67 & 20.61 &  5.99 & 12.09 & +0.82 \\
14 & stack-pytest-withtests         & 18.33 & 18.43 &  3.75 & 13.50 & +0.70 \\
15 & ghactions                      &  8.00 & 17.58 &  7.12 & 10.90 & +0.61 \\
16 & stack-selfdoc-gpt5mini         & 11.00 & 17.79 &  5.99 & 11.59 & +0.61 \\
17 & self-instruct-naive            & 10.00 & 14.70 &  7.87 & 10.86 & +0.57 \\
18 & stack-ruby                     &  9.67 & 16.04 &  7.12 & 10.94 & +0.55 \\
19 & nemotron-junit                 &  7.33 & 15.90 &  7.87 & 10.37 & +0.54 \\
20 & synatra                        &  7.00 & 14.11 &  8.99 & 10.03 & +0.54 \\
21 & stack-selfdoc                  & 14.33 & 15.38 &  5.62 & 11.78 & +0.49 \\
22 & manybugs                       & 18.33 & 12.37 &  5.99 & 12.23 & +0.48 \\
23 & pr-mining                      & 11.00 & 16.58 &  5.99 & 11.19 & +0.48 \\
24 & nebius-swe-agent               & 19.33 & 12.85 &  5.24 & 12.47 & +0.45 \\
25 & stack-pytest                   & 11.00 & 16.70 &  5.62 & 11.11 & +0.43 \\
26 & stack-bash-withtests           & 10.67 & 16.78 &  5.62 & 11.02 & +0.42 \\
27 & go-browse-wa                   & 12.00 & 15.97 &  5.62 & 11.20 & +0.41 \\
28 & exercism-python                &  7.33 & 12.52 &  8.99 &  9.61 & +0.39 \\
29 & softwareheritage               &  9.00 & 16.82 &  5.99 & 10.60 & +0.39 \\
30 & stack-cpp                      & 11.00 & 17.04 &  5.24 & 11.09 & +0.39 \\
31 & stack-junit                    &  8.67 & 14.91 &  7.12 & 10.23 & +0.38 \\
32 & github-dockerfiles             &  9.33 & 12.41 &  8.24 &  9.99 & +0.36 \\
33 & stack-rust                     &  9.33 & 19.68 &  4.12 & 11.04 & +0.36 \\
34 & freelancer                     &  8.67 & 17.07 &  5.62 & 10.45 & +0.33 \\
35 & swegym-openhands               & 20.00 & 13.16 &  4.12 & 12.43 & +0.32 \\
36 & nl2bash                        &  2.00 & 12.88 &  7.12 & 10.00 & +0.31 \\
37 & stack-pytest-gpt5mini          &  8.00 & 16.37 &  5.99 & 10.12 & +0.28 \\
38 & stack-dockerfile               &  8.00 & 17.56 &  5.24 & 10.27 & +0.27 \\
39 & stackexchange-overflow         &  9.67 & 15.89 &  5.62 & 10.39 & +0.27 \\
40 & bugsinpy                       & 12.33 & 15.44 &  4.87 & 10.88 & +0.24 \\
41 & stack-rspec                    &  8.00 & 17.72 &  4.87 & 10.20 & +0.22 \\
42 & stackexchange-codereview       & 11.00 & 14.50 &  5.62 & 10.37 & +0.20 \\
43 & glaive-code-assistant          & 10.33 & 16.81 &  4.49 & 10.54 & +0.19 \\
44 & stack-pytest-synthetic-gpt5nano & 10.67 & 16.53 &  4.49 & 10.56 & +0.19 \\
45 & mind2web                       &  8.67 & 15.32 &  5.62 &  9.87 & +0.15 \\
46 & stack-jest                     &  9.33 & 14.99 &  5.62 &  9.98 & +0.15 \\
47 & stack-bash                     & 11.33 & 12.39 &  6.37 & 10.03 & +0.14 \\
48 & bash-textbook                  &  6.67 & 16.72 &  5.24 &  9.54 & +0.11 \\
49 & stack-phpunit                  &  7.67 & 16.76 &  4.87 &  9.77 & +0.10 \\
50 & stack-csharp                   &  8.67 & 16.12 &  4.87 &  9.89 & +0.09 \\
51 & nemotron-bash-withtests-gpt5mini &  9.33 & 14.52 & 5.24 & 9.70 & +0.04 \\
52 & nemotron-rspec                 &  9.00 & 15.35 &  4.87 &  9.74 & +0.03 \\
53 & nnetnav-live                   & 11.33 & 13.35 &  5.24 &  9.97 & +0.03 \\
54 & bugswarm                       & 13.67 & 11.24 &  5.62 & 10.18 & +0.02 \\
55 & crosscodeeval-csharp           &  5.67 & 13.75 &  6.74 &  8.72 & +0.01 \\
56 & crosscodeeval-python           &  8.00 & 14.30 &  5.62 &  9.31 &  0.00 \\
57 & curriculum-hard                & 10.33 & 15.83 &  3.75 &  9.97 & -0.04 \\
58 & nemotron-bash-withtests        &  5.67 & 15.22 &  5.62 &  8.84 & -0.04 \\
59 & wizardlm-orca                  &  7.00 & 15.73 &  4.87 &  9.20 & -0.04 \\
60 & curriculum-easy                & 10.33 & 14.22 &  4.49 &  9.68 & -0.07 \\
\bottomrule
\end{tabular}
\vspace{10pt}
\caption{\textbf{Full task generation strategy ranking (Part 1 of 2): ranks 1--60.} Continued in \Cref{tab:task_gen_full_part2}. Per-benchmark cells: raw accuracy (\%). Average columns show raw and normalized $z$-score averages.}
\label{tab:task_gen_full_part1}
\end{table}
\FloatBarrier

\begin{table}[!htbp]
\centering
\footnotesize
\setlength{\tabcolsep}{4pt}
\begin{tabular}{rlccccc}
\toprule
 & & \multicolumn{3}{c}{Benchmarks (\%)} & \multicolumn{2}{c}{Average} \\
\cmidrule(lr){3-5} \cmidrule(lr){6-7}
Rank & Strategy & SWE-Bench Verified (100) & OT-TBLite & Terminal-Bench 2.0 & Raw & Normalized \\
\midrule
61 & nemotron-csharp                &  9.30 & 12.70 &  5.62 &  9.21 & -0.08 \\
62 & curriculum-medium              &  8.67 & 13.50 &  5.24 &  9.14 & -0.11 \\
63 & nemotron-bash                  &  7.33 & 11.43 &  6.74 &  8.50 & -0.13 \\
64 & stack-bash-withtests-gpt5mini  &  7.00 & 14.97 &  4.49 &  8.82 & -0.19 \\
65 & crosscodeeval-typescript       &  6.67 & 14.90 &  4.49 &  8.69 & -0.22 \\
66 & defects4j                      &  9.67 & 11.85 &  4.87 &  8.80 & -0.29 \\
67 & nemotron-cpp                   &  5.67 & 14.73 &  4.49 &  8.30 & -0.30 \\
68 & agenttuning-alfworld           &  4.00 & 12.45 &  5.62 &  7.36 & -0.42 \\
69 & qasper                         &  8.00 & 11.51 &  4.87 &  8.13 & -0.42 \\
70 & codeactinstruct                &  4.67 & 11.21 &  5.62 &  7.17 & -0.51 \\
71 & inferredbugs                   & 14.00 &  9.83 &  3.37 &  9.07 & -0.51 \\
72 & crosscodeeval-java             &  6.33 & 12.81 &  4.12 &  7.75 & -0.52 \\
73 & nemotron-rust                  &  4.70 & 11.86 &  5.00 &  7.19 & -0.56 \\
74 & taco                           &  6.00 & 11.06 &  4.87 &  7.31 & -0.58 \\
75 & magicoder                      &  6.00 & 13.11 &  3.37 &  7.49 & -0.65 \\
76 & toolscale                      &  4.40 & 15.26 &  2.62 &  7.43 & -0.65 \\
77 & orca-agentinstruct             &  5.00 & 13.18 &  3.37 &  7.18 & -0.70 \\
78 & stack-go                       &  7.33 & 12.42 &  3.00 &  7.58 & -0.71 \\
79 & codenet-python                 &  4.67 & 10.22 &  4.87 &  6.59 & -0.75 \\
80 & nemotron-pytest                &  5.33 &  8.39 &  4.87 &  6.20 & -0.90 \\
81 & codeforces                     &  3.67 &  8.61 &  5.24 &  5.84 & -0.91 \\
82 & e2egit                         &  2.00 & 10.75 &  4.49 &  5.75 & -0.92 \\
83 & nemo-prism-math                &  2.67 & 11.73 &  3.75 &  6.05 & -0.92 \\
84 & codeelo                        &  3.33 & 10.86 &  3.75 &  5.98 & -0.97 \\
85 & pymethods2test                 &  6.33 &  9.68 &  3.00 &  6.34 & -1.05 \\
86 & quixbugs                       &  5.67 & 12.94 &  1.30 &  6.64 & -1.06 \\
87 & unitsyn-python                 &  3.30 & 10.68 &  3.37 &  5.78 & -1.06 \\
88 & agenttuning-kg                 &  4.00 &  7.08 &  4.49 &  5.19 & -1.18 \\
89 & all-puzzles                    &  2.33 &  8.45 &  3.75 &  4.84 & -1.27 \\
90 & code-contests                  &  2.33 & 11.09 &  2.25 &  5.22 & -1.27 \\
91 & agenttuning-webshop            &  1.67 &  9.14 &  3.37 &  4.73 & -1.31 \\
92 & tulu3-sft-personas-math        &  3.00 &  9.30 &  2.62 &  4.97 & -1.35 \\
93 & agenttuning-db                 &  3.67 &  7.53 &  1.50 &  4.23 & -1.70 \\
94 & agenttuning-mind2web           &  0.33 &  4.65 &  1.12 &  2.03 & -2.26 \\
95 & agenttuning-os                 &  0.00 &  5.64 &  0.37 &  2.00 & -2.31 \\
\bottomrule
\end{tabular}
\vspace{10pt}
\caption{\textbf{Full task generation strategy ranking (Part 2 of 2): ranks 61--95.} Continued from \Cref{tab:task_gen_full_part1}.}
\label{tab:task_gen_full_part2}
\end{table}
\FloatBarrier

\subsection{Mixing Strategies (Full Results)}
\label{app:mixing-full}

\Cref{tab:mixing_full} reports both presentation methods (random shuffle within task and sequential round-robin) for all values of $N$. The random-shuffle Top-4 result is reproduced in \Cref{tab:mixing_strategies} of the main text.

\begin{table}[!htbp]
\centering
\resizebox{\textwidth}{!}{%
\begin{tabular}{rlcccccc}
\toprule
 & & \multicolumn{3}{c}{Benchmarks} & \multicolumn{2}{c}{Average} \\
\cmidrule(lr){3-5} \cmidrule(lr){6-7}
Rank & Mixing Strategy & SWE-Bench Verified (100) & OT-TBLite & Terminal-Bench 2.0 & Raw & Normalized \\
\midrule
\multicolumn{7}{l}{\textit{Random shuffle within task}} \\
\midrule
1 & Top 4 & \textbf{29.33}\,\subscriptval{1.63} & \textbf{17.00}\,\subscriptval{1.71} & \textbf{8.24}\,\subscriptval{1.24} & \textbf{18.19} & \textbf{+0.49} \\
2 & Top 2 & \textbf{29.00}\,\subscriptval{1.60} & \textbf{18.12}\,\subscriptval{1.72} & \textbf{7.12}\,\subscriptval{1.06} & \textbf{18.08} & \textbf{+0.48} \\
3 & Top 8 & 28.00\,\subscriptval{1.70} & 15.86\,\subscriptval{1.70} & \textbf{8.61}\,\subscriptval{1.24} & \textbf{17.49} & \textbf{+0.19} \\
4 & Top 16 & 26.67\,\subscriptval{1.49} & \textbf{19.19}\,\subscriptval{1.70} & 4.12\,\subscriptval{1.06} & 16.66 & -0.10 \\
5 & Top 32 & 20.33\,\subscriptval{1.67} & \textbf{18.67}\,\subscriptval{1.72} & 5.99\,\subscriptval{1.06} & 15.00 & -0.48 \\
6 & Top 1 & \textbf{30.67}\,\subscriptval{1.67} & 14.80\,\subscriptval{1.40} & 4.49\,\subscriptval{0.84} & 16.65 & -0.57 \\
\midrule
\multicolumn{7}{l}{\textit{Sequential round-robin}} \\
\midrule
1 & Top 8 & \textbf{32.67}\,\subscriptval{1.86} & 16.28\,\subscriptval{1.58} & \textbf{8.99}\,\subscriptval{1.40} & \textbf{19.31} & \textbf{+0.46} \\
2 & Top 4 & 28.00\,\subscriptval{1.76} & 17.86\,\subscriptval{1.60} & \textbf{8.61}\,\subscriptval{1.40} & \textbf{18.16} & \textbf{+0.28} \\
3 & Top 2 & 28.67\,\subscriptval{1.70} & \textbf{18.05}\,\subscriptval{1.60} & \textbf{7.49}\,\subscriptval{1.24} & \textbf{18.07} & \textbf{+0.11} \\
4 & Top 32 & 23.00\,\subscriptval{1.76} & \textbf{20.21}\,\subscriptval{1.64} & 6.37\,\subscriptval{1.18} & 16.53 & -0.23 \\
5 & Top 16 & 27.33\,\subscriptval{1.83} & 17.68\,\subscriptval{1.55} & 5.62\,\subscriptval{1.18} & 16.88 & -0.62 \\
\bottomrule
\end{tabular}}
\vspace{10pt}
\caption{\textbf{Full mixing strategy results.} We sweep $N \in \{1, 2, 4, 8, 16, 32\}$, sampling $10{,}000/N$ tasks from each of the top-$N$ sources. Both random shuffling within a training batch and sequential round-robin presentation are evaluated. Mixing the top-4 to top-8 strategies yields the strongest balanced performance under both presentation methods. Per-benchmark cells: raw accuracy (\%) with standard error as subscript. Bolded values are within 1 SE of the column-best.}
\label{tab:mixing_full}
\end{table}
\FloatBarrier

\subsection{Filtering for Longer Episodes: A Compute-Controlled Ablation}
\label{app:compute-controlled-filtering}

\Cref{sec:sft-pipeline} shows that keeping execution traces with more model turns improves the resulting training set. Since longer ($\geq 5$-turn) traces also pack more tokens into each example, the gain could in principle come from one of two sources: the higher-quality multi-turn supervision we intend to select for, or simply a larger training-compute budget. At a fixed row count, the min-turns filter sees roughly 45\% more tokens than an unfiltered subset, so the two explanations are confounded unless we equalize the token budget. We do so directly. We build two training sets that consume the same $\sim$145M tokens: the first keeps the longest ($\geq 5$-turn) episodes ($9{,}859$ rows), and the second draws a random subsample of the pool ($14{,}470$ rows). Everything else about the two runs is identical. \Cref{tab:compute_controlled_filtering} reports the result.

With the token budget held fixed, the min-turns filter still beats the random control by $+3.5$pp on average, with $+5.4$pp on SWE-Bench Verified-100 and $+3.8$pp on Terminal-Bench~2.0 while OT-TBLite moves within the per-benchmark noise. The effect therefore cannot be attributed to extra training compute and longer episodes lead to higher accuracy in multi-turn agentic tasks.

\begin{table}[!t]
\centering
\resizebox{\textwidth}{!}{%
\begin{tabular}{lccccc}
\toprule
 & & \multicolumn{3}{c}{Benchmarks} & Average \\
\cmidrule(lr){3-5} \cmidrule(lr){6-6}
Selection Strategy & Tokens & SWE-bench Verified (100) & OT-TBLite & Terminal-Bench 2.0 & Raw \\
\midrule
Min turns $\geq 5$ (filtered) & 144.76M & \textbf{24.70}\,\subscriptval{1.56} & \textbf{18.40}\,\subscriptval{1.66} & \textbf{10.90}\,\subscriptval{1.35} & \textbf{18.00} \\
Random subsample (control) & 144.78M & 19.30\,\subscriptval{1.63} & 17.10\,\subscriptval{1.48} & 7.10\,\subscriptval{1.24} & 14.50 \\
\bottomrule
\end{tabular}}
\vspace{10pt}
\caption{\textbf{The min-turns filter helps even at a matched token budget.} Both models are Qwen3-8B fine-tuned on subsets, each truncated to an equal $\sim$145M-token budget. Selecting the longest ($\geq 5$-turn) episodes outperforms a random subsample by $+5.4$pp on SWE-bench Verified-100 and $+3.8$pp on Terminal-Bench~2.0, while OT-TBLite is within noise. The benefit therefore reflects genuine multi-turn supervision rather than additional training compute.}
\label{tab:compute_controlled_filtering}
\end{table}

\FloatBarrier

\section{Scaling at 8B}
\label{app:scaling-8b}

\begin{figure}[h]
\centering
\includegraphics[width=\textwidth]{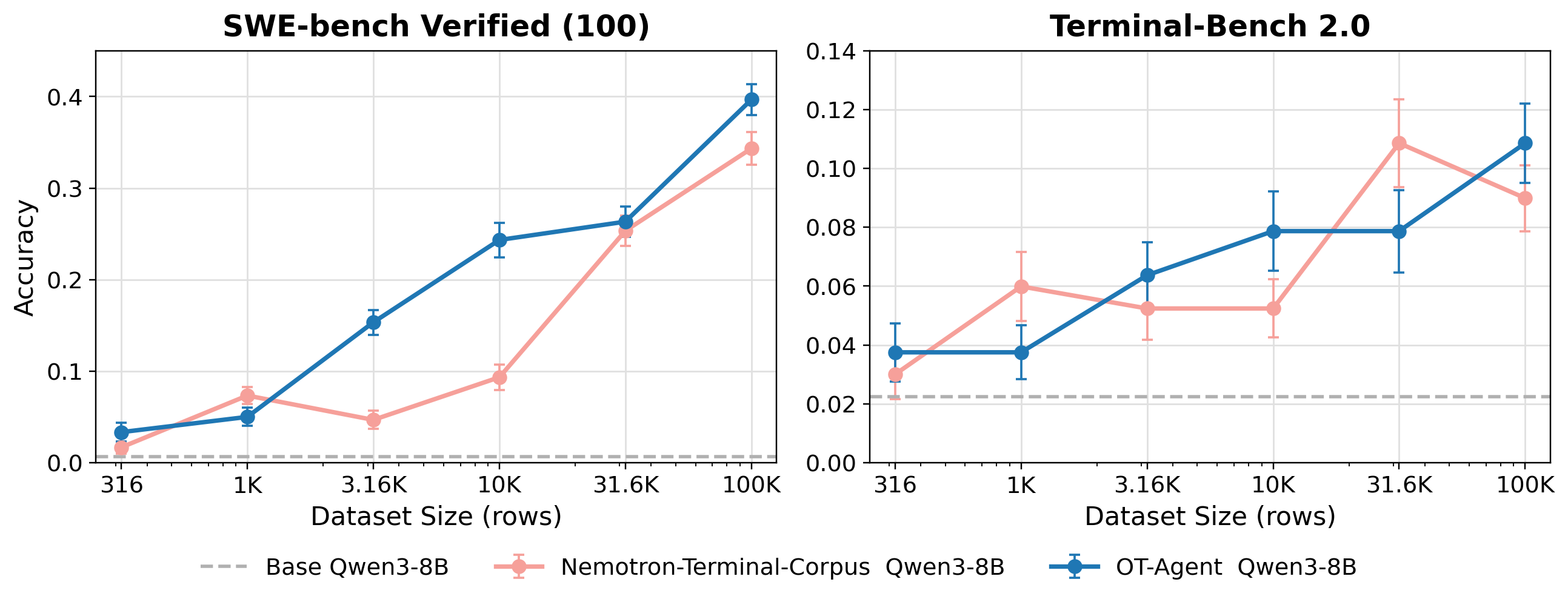}
\caption{\textbf{The OpenThoughts-Agent data recipe scales at 8B.} SWE-bench Verified-100 and Terminal-Bench~2.0 accuracy for Qwen3-8B fine-tuned on OpenThoughts-Agent-v2 across dataset sizes, compared against the Nemotron-Terminal-Corpus baseline and the base Qwen3-8B model (dashed). OpenThoughts-Agent leads at the larger scales and surpasses the baseline on both benchmarks at 100K. Error bars denote standard error across three stochastic re-runs per task.}
\label{fig:scaling-8b}
\end{figure}

Our pipeline ablations (\Cref{sec:sft-pipeline}) use Qwen3-8B as the base model. Here we verify that the final OpenThoughts-Agent data recipe also scales at the 8B model size, mirroring the 32B trends in \Cref{fig:scaling-curves}. \Cref{fig:scaling-8b} reports SWE-bench Verified-100 and Terminal-Bench~2.0 accuracy as we scale the OpenThoughts-Agent-v2 dataset from 316 to 100K rows, alongside the Nemotron-Terminal-Corpus baseline and the base Qwen3-8B model.

OpenThoughts-Agent leads the Nemotron-Terminal-Corpus baseline at most matched dataset sizes on both benchmarks. For example, at 10K rows it reaches 24.3\% versus 9.3\% on SWE-bench Verified-100. As with the 32B model, performance continues to improve at the largest scale rather than plateauing: at 100K rows the 8B reaches 39.7\% on SWE-bench Verified-100 and 10.9\% on Terminal-Bench~2.0 (up from 26.3\% and 7.9\% at 31.6K), surpassing the Nemotron-Terminal-Corpus baseline on both benchmarks (34.3\% and 9.0\%, respectively). This mirrors the effect of synthetic task augmentation at 32B (\Cref{sec:scaling}), where expanding task-description diversity overcomes the upsampling bottleneck.

\FloatBarrier

\section{SFT Hyperparameters}
\label{app:sft-hparams}

\subsection{SFT training hyperparameters per data scale}
\label{appx:sft_hp_per_scale}

\paragraph{Common to all 32B SFT runs.}
All 32B finetunes start from \texttt{Qwen/Qwen3-32B}, use the \texttt{qwen3} (thinking)
chat template, and share the optimization settings below; only the items in
\Cref{tab:sft_hp_per_scale_32b} vary across data scales.

\begin{table}[!htbp]
\centering
\begin{tabular}{ll}
\toprule
Optimizer                          & AdamW, $\beta = (0.9,\,0.98)$, weight decay $0.04$ \\
Learning rate                      & $4\times10^{-5}$ \\
Schedule                           & cosine, warmup ratio $0.1$ \\
Global batch size                  & $96$ \\
Sequence cutoff                    & $32{,}768$ tokens \\
Precision                          & BF16 weights \\
Distributed strategy               & DeepSpeed ZeRO-3 (\texttt{ds\_z3\_accelerate.json}) \\
Hardware                           & $24$ nodes (JUPITER Booster, $4\times$GH200 per node) \\
Shuffle examples between epochs    & disabled (\texttt{disable\_shuffling=True}) \\
Chat template                      & \texttt{qwen3} (thinking) \\
Base model                         & \texttt{Qwen/Qwen3-32B} \\
\bottomrule
\end{tabular}
\vspace{6pt}
\caption{\textbf{Optimizer and infrastructure settings shared across all 32B SFT runs.}}
\label{tab:sft_hp_common_32b}
\end{table}

\paragraph{Per-scale settings (32B).}
The two ablation knobs that vary across data scales are the number of training
epochs and the gradient-clip threshold: smaller scales train for more epochs
with looser clipping, larger scales train for fewer epochs with tighter
clipping (\Cref{tab:sft_hp_per_scale_32b}).

\begin{table}[!htbp]
\centering
\begin{tabular}{lcccl}
\toprule
Data scale & Epochs & Max grad norm & Recipe label & Approx.\ wall time (24 nodes) \\
\midrule
3.16K & 7 & $1\times10^{-4}$ & \texttt{32b\_small}  & $\sim$0.5 h \\
10K   & 7 & $1\times10^{-4}$ & \texttt{32b\_small}  & $\sim$1.5 h \\
31.6K & 5 & $1\times10^{-3}$ & \texttt{32b\_large}  & $\sim$3 h \\
100K  & 5 & $1\times10^{-3}$ & \texttt{32b\_large}  & $\sim$5 h \\
\bottomrule
\end{tabular}
\vspace{6pt}
\caption{\textbf{32B per-scale SFT settings.} The smaller-scale runs (3.16K, 10K) use the
\texttt{32b\_small} recipe (7 epochs, looser gradient clipping); the larger-scale
runs (31.6K, 100K) use \texttt{32b\_large} (5 epochs, tighter clipping). All
other settings (\Cref{tab:sft_hp_common_32b}) are held constant. Wall time is
measured for the diverse-Tezos top-4 family on $24\times$H100 nodes.}
\label{tab:sft_hp_per_scale_32b}
\end{table}

\paragraph{Common to all 8B SFT runs.}
8B runs use the same optimizer and infrastructure family as the 32B runs but
with a non-thinking chat template and a longer training horizon (7 epochs at
all scales). The constant settings are summarized in
\Cref{tab:sft_hp_common_8b}.

\begin{table}[!htbp]
\centering
\begin{tabular}{ll}
\toprule
Optimizer                          & AdamW, $\beta = (0.9,\,0.98)$, weight decay $0.04$ \\
Learning rate                      & $4\times10^{-5}$ \\
Schedule                           & cosine, warmup ratio $0.1$ \\
Global batch size                  & $96$ \\
Sequence cutoff                    & $32{,}768$ tokens (long-context variants: $131{,}072$) \\
Precision                          & BF16 weights \\
Distributed strategy               & DeepSpeed ZeRO-3 \\
Hardware                           & $24$ nodes (JUPITER Booster, $4\times$GH200 per node) \\
Shuffle examples between epochs    & disabled (\texttt{disable\_shuffling=True}) \\
Chat template                      & \texttt{qwen3\_nothink} \\
Base model                         & \texttt{Qwen/Qwen3-8B} \\
\bottomrule
\end{tabular}
\vspace{6pt}
\caption{\textbf{Optimizer and infrastructure settings shared across all 8B SFT runs.}}
\label{tab:sft_hp_common_8b}
\end{table}


\begin{table}[!htbp]
\centering
\begin{tabular}{lccl}
\toprule
Data scale & Epochs & Max grad norm & Approx.\ wall time (24 nodes) \\
\midrule
1K    & 7 & $1.0$ & $\sim$1 h \\
3.16K & 7 & $1.0$ & $\sim$2 h \\
10K   & 7 & $1.0$ & $\sim$4 h \\
31.6K & 7 & $1.0$ & $\sim$10 h \\
100K  & 7 & $1.0$ & $\sim$30 h \\
\bottomrule
\end{tabular}
\vspace{6pt}
\caption{\textbf{8B per-scale SFT settings.} 8B runs hold all hyperparameters
constant across data scales, with only data volume changing. All settings in
\Cref{tab:sft_hp_common_8b} apply.}
\label{tab:sft_hp_per_scale_8b}
\end{table}

\section{RL Hyperparameters and Infrastructure}
\label{app:rl-hparams}

\subsection{Hardware}

\begin{tabular}{ll}
\toprule
Site / scheduler        & NERSC Perlmutter \\
Nodes                   & 6 (2 shared policy / reference, 4 inference) \\
Accelerators            & 4 $\times$ NVIDIA A100-SXM4-80GB per node (24 GPUs total) \\
Interconnect / runtime  & Cray Shasta, Ray cluster, CUDA 12.8, PyTorch bf16 \\
\bottomrule
\end{tabular}

\subsection{Algorithm and optimizer}

\begin{tabular}{ll}
\toprule
RL algorithm                       & RLOO with per-prompt std normalization (\texttt{grpo\_norm\_by\_std=true}) \\
Loss reduction                     & token-mean \\
PPO clip range                     & $[\epsilon_{\text{lo}}, \epsilon_{\text{hi}}] = [0.2, 0.2]$, $c=3$ \\
KL control                         & No \\
Entropy regularization             & No \\
Advantage normalization            & per-batch \\
\midrule
Optimizer (policy)                 & AdamW, $\beta=(0.9,\,0.999)$, weight decay $0$ \\
Optimizer (critic)                 & AdamW, $\beta=(0.9,\,0.999)$, weight decay $0.01$ \\
Learning rate                      & $5 \times 10^{-6}$ \\
Schedule                           & constant (no warmup) \\
Gradient clipping                  & global norm $\le 1.0$ \\
Precision                          & bf16 autocast, fp32 grad accumulation \\
\bottomrule
\end{tabular}

\subsection{Training schedule}

\begin{tabular}{ll}
\toprule
Global steps             & 48 \\
Epochs                   & 2 \\
Train batch (prompts)    & 64 \\
Samples per prompt       & 8 (so 512 trajectories per gradient update) \\
Policy mini-batch        & 64 prompts \\
Update epochs per batch  & 1 \\
Micro-batch (per GPU)    & 1 (train), 4 (forward) \\
Sample packing           & enabled \\
Gradient checkpointing   & enabled (non-reentrant) \\
HF Hub save interval     & every 5 steps \\
\bottomrule
\end{tabular}

\subsection{Distributed strategy}

\begin{tabular}{ll}
\toprule
Strategy                         & FSDP2, \texttt{fsdp\_size=4} (intra-node), CPU param offload \\
Reference / policy colocation    & yes (shared 2-node pool, swapped on/off device) \\
Async generation                 & enabled, max staleness 16 steps, 768 parallel workers \\
Sequence parallel                & 1 (Ulysses backend available, unused) \\
\bottomrule
\end{tabular}

\subsection{Generation (vLLM)}

\begin{tabular}{ll}
\toprule
Engine                          & vLLM async, 16 inference engines, eager mode \\
Sampling                        & temperature 0.7, top-$p$ 0.95, top-$k$ 20 \\
Max generation length           & 4{,}096 tokens \\
Max model context               & 32{,}768 tokens \\
Prefix caching / chunked prefill & enabled \\
KV cache target utilization     & 0.9 of GPU memory \\
Eval sampling                   & greedy ($T=0$, top-$k=-1$), $n=8$ trajectories per prompt \\
\bottomrule
\end{tabular}

\subsection{Rollout environment (Harbor / terminus-2)}

\begin{tabular}{ll}
\toprule
Harbor agent                       & \texttt{terminus-2} with interleaved thinking enabled \\
Per-trial sandbox                  & 1 vCPU, 2{,}048\,MB RAM, 2{,}048\,MB storage \\
Per-trial timeout                  & 1{,}800\,s agent / 120\,s verifier \\
Concurrent trials                  & 280 \\
Number of turns cap per task               & unbounded (effectively gated by timeout) \\
Retry policy                       & 3 retries, exponential backoff (60--600\,s) \\
Failure-mode handling              & masked transient infra errors \\ (Sandbox infra, network) \\ zero reward on \texttt{TimeoutError}, parse errors, OOM \\
\bottomrule
\end{tabular}

\subsection{Headline metrics at step 48}

\begin{tabular}{lr}
\toprule
\texttt{reward/avg\_pass\_at\_8}     & 0.281 \\
\texttt{reward/avg\_raw\_reward}     & 0.107 \\
\texttt{policy/policy\_entropy}      & 0.072 \\
\texttt{policy/raw\_grad\_norm}      & 0.018 \\
Runtime                              & $1.66\times 10^{5}$\,s ($\approx$46\,h) \\
\bottomrule
\end{tabular}

\subsection{Reproducibility Artifacts}
Pinned source commits (as of run start, 2026-02-27):

\begin{tabular}{lll}
\toprule
Repo & Branch & Commit \\
\midrule
\href{https://github.com/open-thoughts/OpenThoughts-Agent}{open-thoughts/OpenThoughts-Agent} & main                      & \texttt{4e2b8422} \\
\href{https://github.com/penfever/SkyRL}{penfever/SkyRL} (fork)                              & \texttt{penfever/working} & \texttt{ada3bd4f} \\
\href{https://github.com/laude-institute/harbor}{laude-institute/harbor}                    & \texttt{penfever/temp-override} & \texttt{94f358bc} \\
\bottomrule
\end{tabular}

Full wandb run at
\href{https://wandb.ai/dogml/OpenThoughts-Agent/runs/fpqklauc}{\texttt{dogml}}, available on request.
Public HF checkpoint available as laion/rl\_swesmith-fixthink-pymethods2test-45. 
We selected a fork of Llama-Factory from \citep{zheng-etal-2024-llamafactory}, extending it to support ALST long-sequence training \citep{bekman2025arcticlongsequencetraining}. Our reinforcement learning framework was an extended version of the popular SkyRL framework from \citep{griggs2025skrylv01}; most of the improvements are described in \cite{cao2025skyrl}. We used \cite{Harbor_Framework} for environment, benchmark and harness management.

\section{RL Run-to-Run Reproducibility}
\label{app:rl-repro}

A natural concern for any RL result is how much of the reported improvement is signal versus run-to-run noise in the training pipeline. To probe this, we evaluate three near-replicate RL runs of the \texttt{pymethods2test} experiment. All three start from the same GLM-4.7-distilled SWE-Smith 8B checkpoint and use the same RLOO recipe, environment, and 24$\times$A100 setup described in \Cref{sec:rl-data} and \Cref{app:rl-hparams}; they differ only in minor training choices (the exported step and, for one run, the learning rate). Because the three runs share a configuration up to these small perturbations, the spread of their downstream eval scores is a direct, if conservative, estimate of the reproducibility of the RL pipeline. The headline checkpoint (\texttt{pymethods2test-45}) was additionally evaluated twice on every benchmark, which lets us separate \emph{eval noise} (re-running the same checkpoint) from \emph{training-run noise} (re-running the pipeline). \Cref{tab:rl_reproducibility} summarizes the comparison; per-cell variance combines within-eval binomial sampling with between-eval variance using the mixed estimator defined in the caption.

\begin{table}[!t]
\centering
\resizebox{\textwidth}{!}{%
\begin{tabular}{rlcccccc}
\toprule
 & & \multicolumn{3}{c}{Benchmarks} & \multicolumn{2}{c}{Average} \\
\cmidrule(lr){3-5} \cmidrule(lr){6-7}
Rank & RL Run & SWE-bench Verified (100) & OT-TBLite & Terminal-Bench 2.0 & Raw & Normalized \\
\midrule
1 & \texttt{pymethods2test-45} & \textbf{35.67}\,\subscriptval{1.83} & 16.02\,\subscriptval{1.58} & \textbf{13.48}\,\subscriptval{1.50} & \textbf{21.72} & \textbf{+0.32} \\
2 & \texttt{pymethods2test} (10-step) & \textbf{35.67}\,\subscriptval{1.83} & \textbf{19.30}\,\subscriptval{1.71} & 8.61\,\subscriptval{1.18} & 21.19 & +0.20 \\
3 & \texttt{pymethods2test} (lr5e-6) & 31.33\,\subscriptval{1.63} & 15.36\,\subscriptval{1.57} & \textbf{12.36}\,\subscriptval{1.45} & 19.68 & -0.51 \\
\bottomrule
\end{tabular}}
\vspace{10pt}
\caption{\textbf{Run-to-run reproducibility of the 8B agentic RL pipeline.} Three near-replicate RL runs of the \texttt{pymethods2test} experiment, all starting from the same GLM-4.7-distilled SWE-Smith 8B checkpoint and trained with RLOO on 24$\times$A100, agree; the \texttt{lr5e-6} run uses a learning rate of $5\mathrm{e}{-6}$ rather than the default of the other two.}
\label{tab:rl_reproducibility}
\end{table}

\textbf{Replicate RL runs differ by only $\approx$1.6 points on ID and $\approx$2.0 points on OOD.} The in-distribution (Core) set means lie within $20.2$--$21.8\%$ (range $1.6$ pp, cross-run standard deviation $0.8$ pp) and the out-of-distribution means within $26.5$--$28.5\%$ (range $2.0$ pp, cross-run standard deviation $1.0$ pp), including the learning-rate variant. This run-level spread is comparable to, and slightly larger than, the per-cell eval noise: the two repeated evaluations of \texttt{pymethods2test-45} differ by $0.3$--$3.7$ points on most benchmarks, with the small-$N$ FinanceAgent-Terminal (50 tasks) the main outlier at $\approx$11 points, consistent with its large binomial sampling error. At the benchmark level, larger denominators are correspondingly more stable: SWE-Bench-Verified (500 tasks) varies by only $1.5$ points across runs while the 50--127-task OOD benchmarks vary by $2.7$--$5.3$ points, indicating that most of the per-benchmark variability is eval sampling noise rather than genuine training-run differences. The two single-eval runs (10-step and lr5e-6) are reported with binomial-only error bars, since a single eval cannot estimate run-level variance; their means nonetheless fall inside the band set by the twice-evaluated headline run.

\textbf{Takeaway.} The pipeline is reproducible at the $\sim$2-point level: the gains reported in \Cref{tab:ot_agent_8B_table3} -- the $\approx$5-point RL-specific gain on the Core benchmarks over the SFT-only checkpoint (e.g.\ $+5.4$ on SWE-Bench-Verified), as well as the $\approx$18-point gain of the full SFT+RL pipeline over the Qwen3-8B base -- are substantially larger than the $\approx$1.6/2.0-point ID/OOD run-to-run spread measured here, so they are unlikely to be artifacts of training-run noise. Reporting set-level means with the mixed-variance error above, rather than single best-of-many eval numbers, is the appropriate way to compare RL checkpoints at this scale.

\section{What RL Learns: Emergent Behavior Behind the \texttt{pymethods2test} Result}
\label{app:rl-behavior}

\Cref{tab:rl_data_ablation} establishes \emph{that} \texttt{pymethods2test} is the strongest RL data source, but not \emph{why}. Here, we offer a detailed mechanistic account of the behavioral changes that emerge after RL. We compare the headline \texttt{pymethods2test} run to the \texttt{llm-verifier-freelancer}  run, the strongest OOD alternative in \Cref{tab:rl_data_ablation}). Our analysis combines three views: (i) time-binned behavioral statistics over the RL traces themselves ($\sim$11k hero rollouts, $\sim$53k baseline rollouts); (ii) per-trace behavioral deltas between the pre-RL base checkpoint and the post-RL checkpoint, measured on held-out eval traces; and (iii) a pairwise LLM judge (\texttt{gpt-5-2025-08-07}) that reads $30$ same-task pre/post trace pairs per run and reports a winner, confidence, and behavioral tags.

\subsection{Legitimate exploration, not reward hacking}
\label{app:rl-behavior-hero}

\begin{table}[t!]
\centering
\newcolumntype{C}[1]{>{\centering\arraybackslash}p{#1}}
\resizebox{0.86\textwidth}{!}{
\begin{tabular}{l C{2.2cm} C{2.2cm} C{2.4cm}}
\toprule
Per-trace behavioral feature & Pre-RL & Post-RL & $\Delta$ (rel.) \\
\midrule
think tokens / trace                & $30.3$  & $65.4$  & $+35.1$ ($+116\%$) \\
think blocks / trace                & $0.21$  & $0.44$  & $+0.23$ ($+110\%$) \\
self-correction phrases / trace     & $0.63$  & $1.14$  & $+0.51$ ($+81\%$)  \\
assistant tokens / trace            & $4254$  & $6499$  & $+2245$ ($+53\%$)  \\
tool errors / trace                 & $5.9$   & $8.8$   & $+2.9$ ($+48\%$)   \\
tool responses / trace              & $18.5$  & $24.6$  & $+6.1$ ($+33\%$)   \\
assistant msgs / trace              & $20.0$  & $26.3$  & $+6.2$ ($+31\%$)   \\
tool calls / trace                  & $31.3$  & $40.9$  & $+9.6$ ($+31\%$)   \\
mean tokens / asst.\ msg            & $204.7$ & $243.3$ & $+38.6$ ($+19\%$)  \\
tool error rate                     & $31.6\%$& $35.8\%$& $+4.1$ pp          \\
think / assistant ratio             & $0.5\%$ & $0.6\%$ & $+0.1$ pp          \\
code fences / trace                 & $0.15$  & $0.14$  & $-0.01$ ($-7\%$)   \\
\midrule
\multicolumn{4}{l}{\emph{Macro metrics (whole-trace)}} \\
mean turns / trace                  & $40.5$  & $53.3$  & $+12.9$ ($+32\%$)  \\
mean tokens / conversation          & $18432$ & $23686$ & $+5254$ ($+29\%$)  \\
\midrule
\multicolumn{4}{l}{\emph{Tool-call mix (share of all tool calls)}} \\
\texttt{mark\_task\_complete}       & $1.9\%$ & $3.0\%$ & $+1.1$ pp          \\
\texttt{bash\_command}              & $98.1\%$& $97.0\%$& $-1.1$ pp          \\
\bottomrule
\end{tabular}}
\vspace{4pt}
\caption{\textbf{Behavioral shift induced by \texttt{pymethods2test} RL, measured on held-out SWE-bench-Verified eval traces.}
Per-trace means computed over $300$ eval rows for the pre-RL base checkpoint
(\texttt{...GLM\_4\_7\_swesmith\_sandboxes...}) versus the post-RL checkpoint
(\texttt{...rl\_\_24GPU\_base\_\_exp\_rpt\_pymethods2test...}); the same $100$ underlying
SWE-bench-Verified tasks are run $\times 3$ on each side. Features are sorted within each
block by the magnitude of the (scale-normalized) delta. The dominant change is a large
increase in \emph{exploratory activity}: the agent thinks more, calls more tools, emits
more assistant messages, and self-corrects more. Crucially the per-call \emph{tool error
rate} rises only $+4.1$ pp, so the $+48\%$ rise in absolute tool errors is mostly a
by-product of issuing more calls rather than degraded tool-use competence, and the share of
\texttt{mark\_task\_complete} calls actually rises ($+1.1$ pp) -- evidence against a
trivial reward-hacking or formatting-only explanation. On these same $100$ tasks the policy
flips $18$ from fail to pass against a single regression.}
\label{tab:rl_behavior_shift}
\end{table}

The largest and most consistent shifts are all increases in \emph{exploratory activity}. Reasoning expands sharply -- think tokens per trace more than double ($30.3\!\to\!65.4$, $+116\%$) and think blocks roughly double ($0.21\!\to\!0.44$) -- alongside more self-correction ($0.63\!\to\!1.14$ phrases per trace, $+81\%$), more tool calls ($31.3\!\to\!40.9$, $+31\%$), more assistant messages ($20.0\!\to\!26.3$), and longer conversations ($+12.9$ turns, $+29\%$ tokens). The post-RL policy is thus \emph{more} exploratory than even the already-verbose distilled base. Over $30$ same-task pre/post trace pairs, the judge prefers the post-RL hero policy on $25/30$ ($83.3\%$, $0$ ties, $24/30$ high-confidence). The most frequent behavioral tags are \texttt{more-tool-calls} ($73\%$ of pairs), \texttt{longer-trace} ($53\%$), \texttt{more-tool-errors} ($50\%$), and \texttt{different-solution-strategy} ($47\%$) -- the judge independently sees the same expansion the metrics show. The change is \emph{not} uniformly ``more verbose,'' however: $8/30$ pairs ($27\%$) are tagged \texttt{fewer-tool-calls}/\texttt{shorter-trace}, i.e.\ on a sizable minority of tasks RL taught the policy to \emph{condense}.

Three observations argue that this is genuine capability change rather than a reward-hacking or formatting artifact. First, the post-RL policy emits \emph{more} tool calls \emph{and} more broken tool calls in absolute terms ($5.9\!\to\!8.8$ tool errors per trace, $+48\%$), which is the opposite of what a policy gaming a brittle parser would do. Second, the per-call tool \emph{error rate} rises only $+4.1$ pp ($31.6\%\!\to\!35.8\%$): the extra absolute errors are mostly explained by issuing more calls, not by tool-use competence degrading. Third, the share of \texttt{mark\_task\_complete} calls actually \emph{rises} ($1.9\%\!\to\!3.0\%$), inconsistent with a policy that learned to end early to harvest a formatting reward. The behavioral expansion converts to held-out gains: on the $100$ shared SWE-Bench-Verified tasks, RL flips $18$ tasks from fail to pass against a single regression (\texttt{sympy\_\_sympy-15017}).

\paragraph{Why this data source pushed the policy to explore.}
The RL-time reward trajectory (\Cref{fig:rl_hero_reward}, blue) is the key context: \texttt{pymethods2test} presents a \emph{medium, non-saturated} reward (it hovers around $0.47$--$0.51$, far from the ceiling) that is \emph{hard to improve}. Under RLOO this is exactly the regime that rewards trying harder -- thinking more, calling more tools, attempting more fixes -- because incremental returns come from working a problem more thoroughly rather than from a quick exploit. Across the run, mean conversation length and think budget grow while the reward plateaus, until the policy over-extends. We read the collapse as the downside of the same exploration pressure that produced the gains, not as a separate failure: the reproducibility study in \Cref{app:rl-repro} (which exports at step $45$, and a $10$-step and an \texttt{lr5e-6} variant) is the corresponding evidence that the \emph{pre-collapse} regime is stable to re-run.

\subsection{The contrast: the LLM-verifier run compacts instead of explores}
\label{app:rl-behavior-baseline}

\begin{table}[t!]
\centering
\newcolumntype{C}[1]{>{\centering\arraybackslash}p{#1}}
\resizebox{0.96\textwidth}{!}{
\begin{tabular}{l C{4.2cm} C{4.2cm}}
\toprule
Axis (pre-RL $\to$ post-RL) & \texttt{pymethods2test} (hero) & \texttt{llm-verifier-freelancer} (baseline) \\
\midrule
RL-time reward trajectory      & $0.47 \to 0.51$ peak $\to 0.14$ collapse & $0.54 \to 0.73$ near-monotonic \\
tail-bin agent-timeout rate    & $\approx 80\%$                  & stayed $18$--$36\%$ throughout \\
completed-trial reward (errors excl.)\textsuperscript{$\ddagger$} & $0.652 \to 0.541$ ($-0.11$)  & $0.415 \to 0.493$ ($+0.08$)  \\
mean turns / trace             & $+12.9$ ($+32\%$)               & $-7.8$ ($-11\%$)             \\
mean tokens / conversation     & $+29\%$                         & $+1\%$ (flat)               \\
tool calls / trace             & $+31\%$                         & $-8\%$                      \\
think tokens / trace           & $+116\%$                        & $-42\%$                     \\
self-correction phrases        & $+81\%$                         & $-59\%$                     \\
LLM-judge post-RL win rate     & $83.3\%$ ($25/30$)              & $73.3\%$ ($22/30$)          \\
dominant LLM-judge tag         & \texttt{more-tool-calls} ($73\%$) & \texttt{fewer-tool-calls} ($53\%$) \\
\bottomrule
\end{tabular}}
\vspace{4pt}
\caption{\textbf{Two RL runs on the same pipeline learn opposite behavioral policies.}
The \texttt{pymethods2test} ``hero'' run and the \texttt{llm-verifier-freelancer}
``baseline'' run share the identical pipeline (GLM-4.7-distilled SWE-Smith 8B base, RLOO
recipe, rollout environment, $24\times$A100), differing only in the RL data source.
On \emph{every} behavioral axis the two runs move in opposite directions: the hero policy
\emph{expands} (more turns, tokens, tool calls, thinking, self-correction) while the
baseline policy \emph{compacts} (fewer turns, calls, thinking, self-correction). The
hero's expansion coincides with a non-monotonic reward curve that peaks near step $\sim$35
and then collapses (\Cref{fig:rl_hero_reward}); the baseline's compaction coincides with a
smooth, monotone reward rise (\Cref{fig:rl_baseline_reward}). We read this as
exploration-purchased-at-the-cost-of-stability (hero) versus tightening-of-an-existing
strategy (baseline). RL-time reward and behavioral deltas are from the time-binned trace
analysis and the pre/post eval-trace deltas; LLM-judge win rates are pairwise
\texttt{gpt-5-2025-08-07} comparisons over $30$ same-task trace pairs per run.
\textsuperscript{$\ddagger$}\,Mean reward over trials that produced a reward (errored/timed-out trials excluded) -- a \emph{conditional} reward, distinct from the all-trial Harbor Mean accuracy (errors$=0$) reported in \Cref{tab:ot_agent_8B_table3,tab:rl_data_ablation}. The hero's conditional reward dips here while its all-trial accuracy roughly doubles ($0.19\!\to\!0.33$); see \Cref{app:rl-behavior-downstream}.}
\label{tab:rl_behavior_hero_vs_baseline}
\end{table}

Running the identical pipeline on \texttt{llm-verifier-freelancer} produces the mirror image (\Cref{tab:rl_behavior_hero_vs_baseline}). Its RL-time reward rises \emph{near-monotonically} from $\approx 0.54$ to $\approx 0.73$ with no collapse (\Cref{fig:rl_baseline_reward}), and along the way the policy \emph{compacts}: mean turns fall ($-7.8$), tool calls fall ($-8\%$), think tokens fall ($-42\%$), and self-correction phrases fall sharply ($19.7\!\to\!8.0$ per trace, $-59\%$), while tokens per conversation stay essentially flat ($+1\%$). Every behavioral axis on which the hero run moved \emph{up}, the baseline run moved \emph{down}. The LLM judge sees the same thing from the other direction: it prefers the post-RL baseline policy on $22/30$ pairs ($73.3\%$), but now the dominant tags are \texttt{fewer-tool-calls} ($53\%$), \texttt{shorter-trace} ($40\%$), and \texttt{different-tool-strategy} ($40\%$). Its top judgments describe a pre-RL policy trapped in loops of malformed tool arguments that RL taught to stop dithering and ship; on \texttt{financeagent-14} (post-RL wins, high confidence) ``the baseline meandered with many turns (113) and tool calls (55) $\dots$ The post-RL trace kept things short (12 turns, 6 tool calls) $\dots$ avoided EDGAR API usage and did not self-correct, proceeding more directly toward the target metric.''

\subsection{Connecting behavior to downstream evals}
\label{app:rl-behavior-downstream}

\begin{figure}[t]
  \centering
  \includegraphics[width=0.92\linewidth]{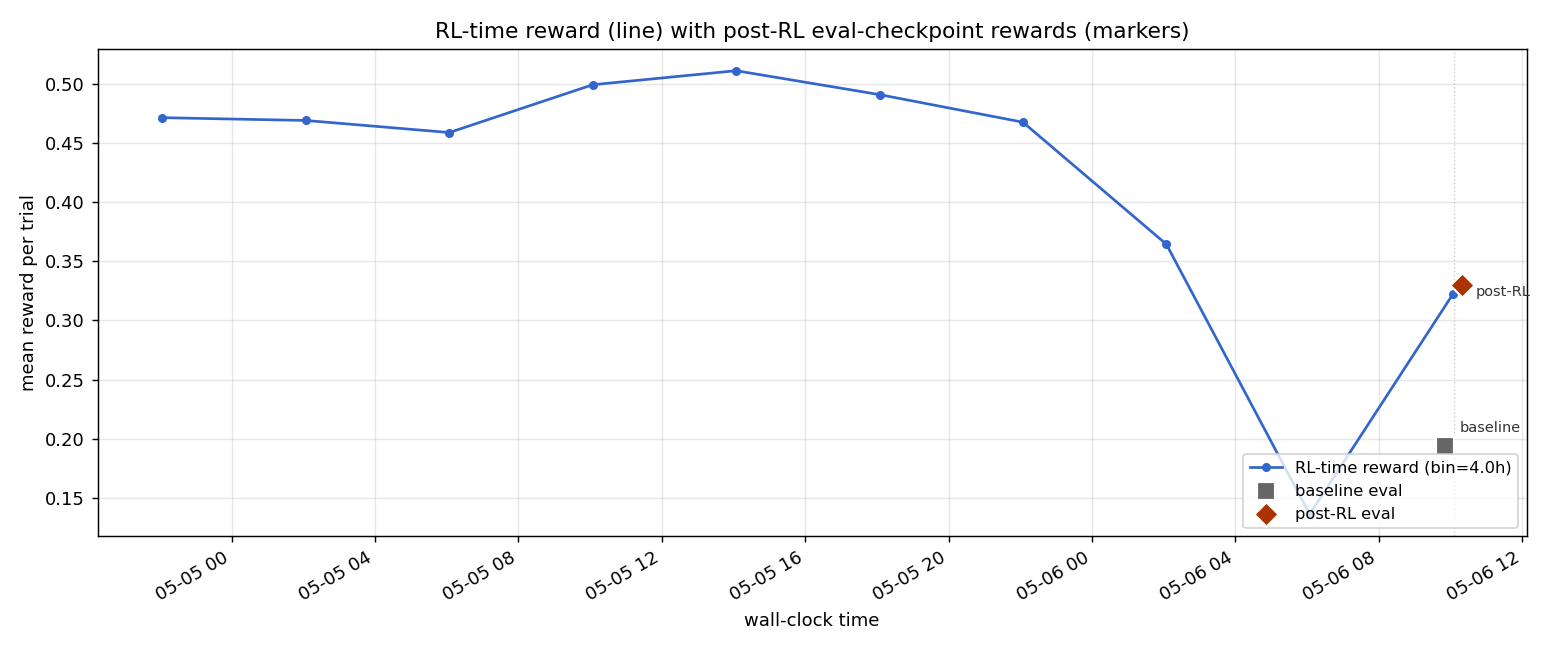}
  \caption{\textbf{Hero (\texttt{pymethods2test}) RL-time reward peaks and then collapses.}
  Mean reward per rollout in $4$-hour wall-clock bins (blue line) over the hero run, with
  the post-RL and pre-RL eval-checkpoint mean rewards on held-out SWE-Bench-Verified marked
  on the right (diamond $=$ post-RL, square $=$ pre-RL base). Reward rises modestly to a
  peak near $0.51$, then collapses to $\approx 0.13$ as the policy over-explores (mean turns
  and think tokens spike while productive tool calls fall and the agent-timeout rate reaches
  $\approx 80\%$). The deployed checkpoint is taken from \emph{before} this collapse; it
  roughly doubles the base policy's eval reward ($0.33$ vs.\ $0.19$).}
  \label{fig:rl_hero_reward}
\end{figure}

\begin{figure}[t]
  \centering
  \includegraphics[width=0.92\linewidth]{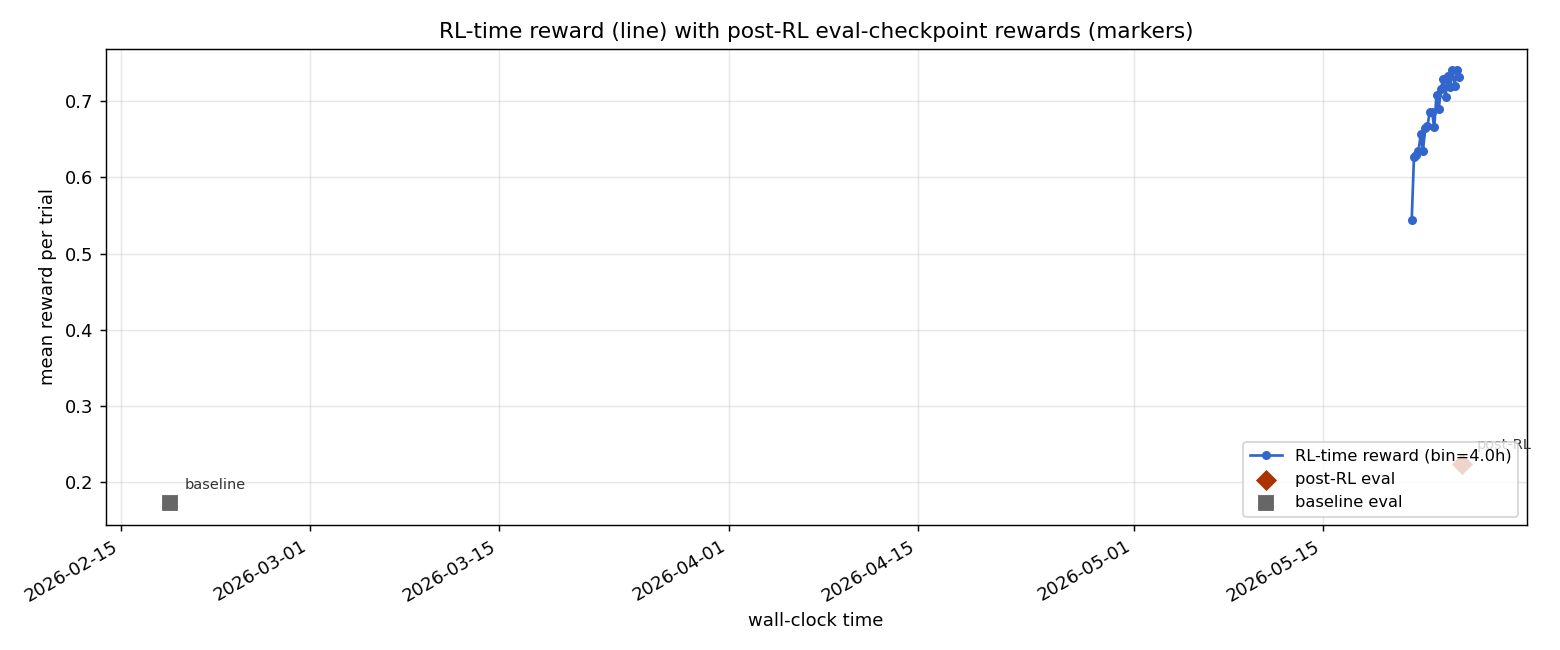}
  \caption{\textbf{Baseline (\texttt{llm-verifier-freelancer}) RL-time reward rises near-monotonically.}
  Mean reward per rollout in $4$-hour bins (blue line, clustered at right of the wall-clock
  axis because the pre-RL base eval -- gray square -- predates training by months) rises
  smoothly from $\approx 0.54$ to $\approx 0.73$ with no collapse, the mirror image of
  \Cref{fig:rl_hero_reward}. The corresponding policy \emph{compacts} (fewer turns, tool
  calls, and think tokens; see \Cref{tab:rl_behavior_hero_vs_baseline}) rather than expands.}
  \label{fig:rl_baseline_reward}
\end{figure}

The behavioral changes persist into held-out evaluation and track the downstream ranking in \Cref{tab:rl_data_ablation}. On the $300$-row SWE-Bench-Verified eval, the pre-collapse hero checkpoint roughly \emph{doubles} the base policy's \emph{all-trial} reward -- the Harbor Mean accuracy that counts errored/timed-out trials as $0$, the same metric reported in \Cref{tab:ot_agent_8B_table3,tab:rl_data_ablation} (post-RL $0.33$ vs.\ pre-RL $0.19$; \Cref{fig:rl_hero_reward}, markers) -- and flips $18$ tasks fail$\to$pass. This all-trial accuracy rises \emph{even though} the mean reward over \emph{completed} trials dips ($0.652\!\to\!0.541$, \Cref{tab:rl_behavior_hero_vs_baseline}): RL converts many trials the base used to abandon to timeouts into completed attempts, and because those rescued trials are the harder ones, the conditional average falls while the all-trial accuracy climbs The baseline's compaction also helps its own evals (post-RL $0.224$ vs.\ pre-RL $0.173$ on the $156$-row FinanceAgent eval), which is why \texttt{llm-verifier-freelancer} is competitive on OOD in \Cref{tab:rl_data_ablation}; but compaction transfers less well to ID Core benchmarks that reward thoroughness.

\begin{figure}[t]
  \centering
  \includegraphics[width=\linewidth]{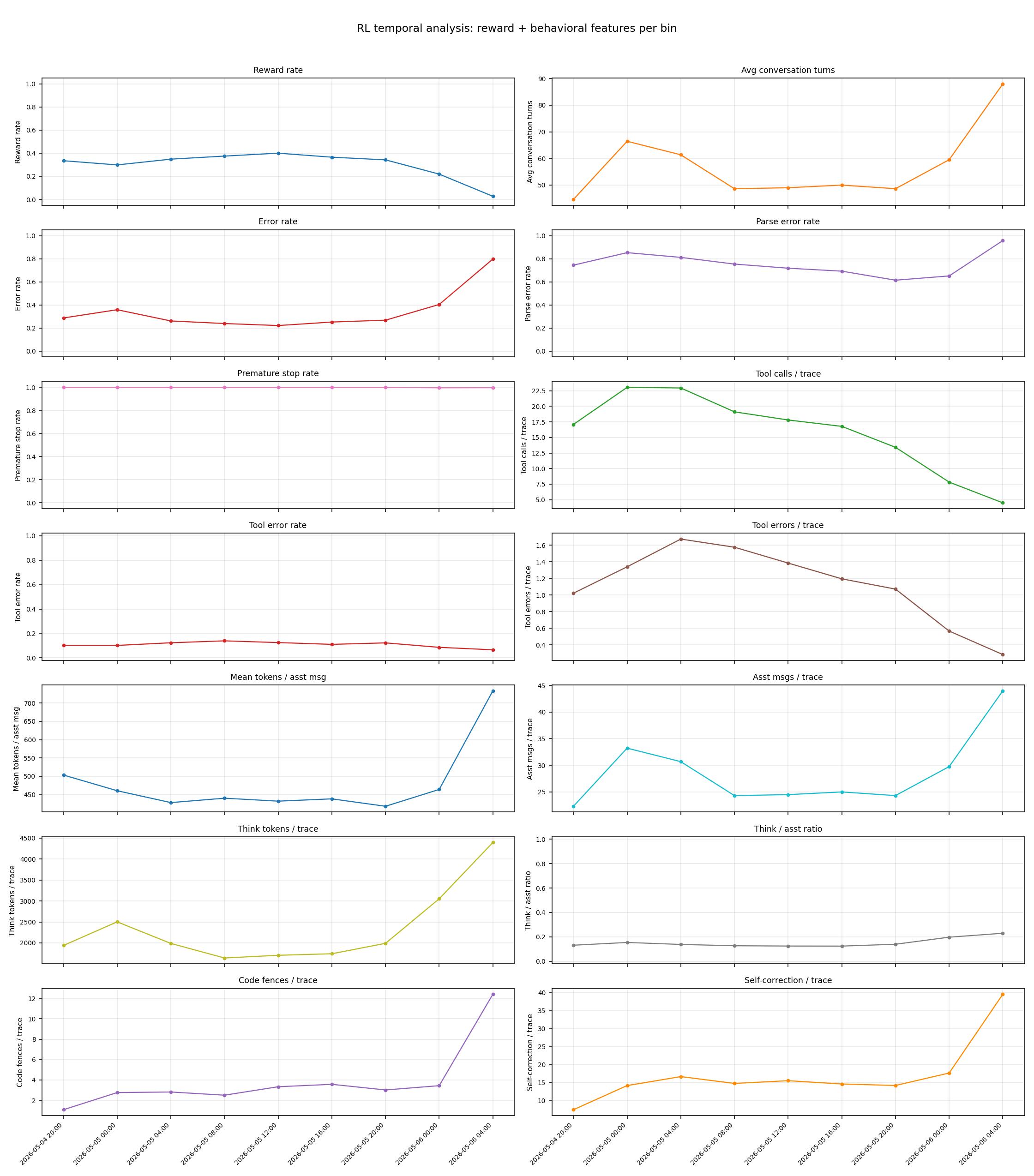}
  \caption{\textbf{Time-binned behavioral dynamics of the hero (\texttt{pymethods2test}) run.}
  Reward, error/parse-error/premature-stop rates, and nine per-trace behavioral features
  over the run ($4$-hour bins on a shared RL-time axis). The exploration-then-collapse
  signature is visible across panels: as reward plateaus and then falls, mean conversation
  turns, mean tokens per assistant message, think tokens per trace, and self-correction per
  trace all spike in the final bins, while tool calls per trace and tool errors per trace
  fall -- the policy substitutes long internal reasoning for productive tool use and times
  out. This is the run-level analogue of the per-task over-exploration the LLM judge flags
  on tasks like \texttt{django\_\_django-13128}.}
  \label{fig:rl_hero_temporal}
\end{figure}

\section{Eval Setup}
\label{app:eval}


\paragraph{Evaluation Stack Overview} For evaluation, we serve models using vLLM \citep{kwon2023efficient}, orchestrate agent trials through Harbor \citep{Harbor_Framework}, and isolate each trial in a Daytona \citep{Daytona_Infrastructure_2025} sandbox. We evaluate on the following benchmarks: 

\begin{itemize}
  \item \textbf{Terminal Bench 2.0} (89 tasks)~\citep{merrill2026terminalbenchbenchmarkingagentshard}:
        Hand-crafted, human-verified tasks that spans SWE, biology, security, system administration, and machine learning.
        
  \item \textbf{SWE-Bench-Verified-100} (100 tasks)~\citep{jimenez2024swebenchlanguagemodelsresolve}:
        a stratified subsample (by repository) of the 500-task human-validated SWE-Bench Verified
        split where agent produces a patch against a Python repo at a fixed commit.

  \item \textbf{SWE-Bench-Verified} (500 tasks)~\citep{jimenez2024swebenchlanguagemodelsresolve}:
        full SWE-Bench Verified split, same format as above.

  \item \textbf{OpenThoughts-TBLite} (100 tasks)~\citep{OpenThoughts-TBLite}:
        a curated collection of 100 Terminal-Bench style tasks that balanced across four
        difficulty buckets and engineered as a fast proxy for the full Terminal-Bench 2.0 performance;

  \item \textbf{Aider Polyglot} (225 tasks)~\citep{aiderpolyglot2024}:
        A multi-language code-editing benchmark (C++, Go, Java, JavaScript,
        Python, Rust) sourced from Exercism and filtered to the harder
        problems.

  \item \textbf{BFCL-Parity} (123 tasks)~\citep{patil2025bfcl}:
        Berkeley Function Calling Leaderboard (BFCL) is a comprehensive benchmark for evaluating large language models' ability to call functions/tools correctly based on natural language instructions. This is a stratified random subsample of BFCL v4.

  \item \textbf{MedAgentBench} (300 tasks)~\citep{jiang2025medagentbench}:
        A clinical-agent benchmark over a vendored FHIR server populated
        with $\sim$700k records.

  \item \textbf{GAIA-127} (127 tasks)~\citep{mialon2023gaia}:
        A text-only subset of the GAIA validation split for general-AI-assistant
        multi-step factoid QA with web browsing, file handling, and tool
        use. For this benchmark, we evaluated in a default setting without setting a search API.

  \item \textbf{FinanceAgent-Terminal} (50 tasks)~\citep{valsai2024financeagent}:
        A terminal-agent variant of Vals AI's Finance Agent Benchmark
        over SEC 10-K/10-Q filings, with shell access to EDGAR/web/RAG
        tools. For this benchmark, we evaluated while providing SERP and EDGAR APIs available to the models.
\end{itemize}

All datasets are prepared in Harbor format, and for OOD benchmarks we generated the dataset using Harbor adapter~\citep{Harbor_Framework}.

\paragraph{Evaluation Setup} 
\label{app:eval-setup}
For every (model, benchmark, harness) combination, we performed 3 repeated runs and report average pass@1 accuracy and standard error across runs. All evaluations with our trained model run with a 32K context window, a 16K output cap, and a proactive summarization threshold of 2048 tokens. Each fire defaults to 32 concurrent trials, with Daytona snapshot registered and reused whenever available to avoid environment rebuilding during repeated evaluations. This mitigate the occurrence of transient environment build errors and speed up evaluations overall. For every baseline model we report, we run the above procedure twice: once under \texttt{terminus-2} (parity with the setup of our trained model) and once under the \emph{preferred scaffold and serving configuration} indicated from the original report. 

\paragraph{Timeout-aware Evaluation}
\label{app:eval-time}
Agentic trials are token-bound: a trial that takes
ten minutes for an 8B model decoding at $\sim$150\,tokens/s may take
an hour for a 32B model decoding at $\sim$30\,tokens/s, with the
\emph{same} number of agent turns. Holding per-trial timeouts
constant across model sizes therefore confounds accuracy --- slower
models time out on tasks faster ones complete. To compensate this, we applied a timeout multiplier $m$ (through the \texttt{-{}-timeout-multiplier} flag in harbor configuration).
We use $m = 2$ for 8B fires and $m = 16$ for 32B fires. To avoid overly time-consuming evaluations, after applying the multipliers we employed a 2-hour cap on agent execution regardless of $m$:
\[
  \text{effective\_timeout}
  \;=\; \min\!\big(\text{cap},\; \text{base} \times m\big).
\]
For larger models where we cannot get comparable throughput, we used third-party API to ensure fair comparison. 

\paragraph{Coderforge-Preview}We do not include in the main results table open-data releases that did not release a model; the most prominent example of this is CoderForge-Preview~\citep{CoderForge2026}.

\paragraph{Reproducing OpenSWE}
\label{app:eval_reprod_openswe}
OpenSWE~\citep{fu2026davincienvopensweenvironment} reports $62.4\%$ on SWE-Bench-Verified (full $500$-task split) training from a Qwen-2.5-32B base. We attempted to reproduce this score using the publicly released \texttt{GAIR/OpenSWE-32B} checkpoint. To match their setup, we ran with the same temperature=0.7, $128\text{K}$-token context window, 300 step budget, and together with a SWE-Agent harness configuration the authors shared with us in correspondence. The OpenSWE repository mainly provides detail for data pipelines, but scaffold or evaluation-harness components were not publicly released at the time of writing, so we cannot guarantee an exact match in every internal detail. On our SWE-Bench-Verified-100 subset we measure $44.0\%$; we do not yet have a directly comparable number on the full $500$-task split, so the gap to the reported $62.4\%$ is partly across different denominators in addition to whatever methodological differences remain. We have not been able to fully attribute the residual gap and continue to investigate. In the meantime, we flag OpenSWE in Table~\ref{tab:appendix_eval_table} and exclude it from main Table~\ref{tab:ot_agent_main_table1}, while still citing the reported number with an underline marker.

\begin{table}[p]
\centering
\newcolumntype{C}[1]{>{\centering\arraybackslash}p{#1}}
\resizebox{\textwidth}{!}{
\begin{tabular}{p{0.15cm} l | C{0.9cm}C{0.9cm}C{0.9cm}C{0.9cm} @{\hspace{0.15cm}}p{0.1cm}@{\hspace{0cm}} C{0.9cm}C{0.9cm}C{0.9cm}C{0.9cm}C{0.9cm} @{\hspace{0.15cm}}p{0.1cm}@{\hspace{0cm}} C{0.65cm}}
\toprule
 & Model & \rotatebox{45}{\textbf{OpenThoughts-TBLite}} & \rotatebox{45}{\textbf{SWE-Bench-Verified-100}} & \rotatebox{45}{\textbf{SWE-Bench-Verified}} & \rotatebox{45}{\textbf{Terminal-Bench 2.0}} & \rotatebox{45}{\scriptsize{------ OOD ------}} & \rotatebox{45}{\textbf{Aider-Polyglot}} & \rotatebox{45}{\textbf{BFCL-Parity}} & \rotatebox{45}{\textbf{MedAgentBench}} & \rotatebox{45}{\textbf{GAIA-127}} & \rotatebox{45}{\textbf{FinanceAgent-Terminal}} & \rotatebox{45}{\scriptsize{------ Avg ------}} & \rotatebox{45}{\textbf{Avg}} \\
\midrule
\multirow{11}{*}{\rotatebox{90}{\textit{~8B Scale}}} & Nemotron-Terminal-8B \citep{pi2026dataengineeringscalingllm} & \textbf{25.8}$_{1.8}$ & 21.7$_{1.7}$ & 22.1$_{0.7}$ & \textbf{13.1}$_{1.4}$ & \multicolumn{1}{!{\vrule width 0.5pt}c}{} & 12.6$_{1.0}$ & 59.9$_{1.6}$ & \textbf{48.6}$_{1.2}$ & \textbf{14.4}$_{1.3}$ & 11.3$_{2.0}$ & \multicolumn{1}{!{\vrule width 0.5pt}c}{} & 26.0 \\
 & SWE-Lego-Qwen3-8B \citep{tao2026swelego} & 3.2\rlap{\textsuperscript{$\ast$}}$_{0.8}$ & \textbf{42.3}\rlap{\textsuperscript{$\ast$}}$_{1.9}$ & \textbf{37.6}\rlap{\textsuperscript{$\ast$}}$_{0.8}$ & 0.7$_{0.4}$ & \multicolumn{1}{!{\vrule width 0.5pt}c}{} & 17.5\rlap{\textsuperscript{$\ast$}}$_{1.0}$ & \textbf{75.3}\rlap{\textsuperscript{$\ast$}}$_{1.1}$ & 5.0$_{0.7}$ & 6.0\rlap{\textsuperscript{$\ast$}}$_{0.8}$ & 1.3$_{0.9}$ & \multicolumn{1}{!{\vrule width 0.5pt}c}{} & 20.5 \\
 & Endless-Terminal-OpenThinker-8B \citep{gandhi2026endlessterminalsscalingrl} & 13.6$_{1.4}$ & 11.3$_{1.5}$ & 12.5$_{0.6}$ & 8.2$_{1.1}$ & \multicolumn{1}{!{\vrule width 0.5pt}c}{} & \textbf{18.8}$_{1.0}$ & 48.8$_{1.6}$ & 18.4$_{1.0}$ & 5.2$_{0.8}$ & 10.0$_{2.1}$ & \multicolumn{1}{!{\vrule width 0.5pt}c}{} & 17.4 \\
 & OpenThinker-Agent-v1 & 10.9$_{1.4}$ & 10.0$_{1.4}$ & 14.0$_{0.6}$ & 3.7$_{1.0}$ & \multicolumn{1}{!{\vrule width 0.5pt}c}{} & 16.7$_{1.1}$ & 40.9$_{1.7}$ & 20.0$_{1.0}$ & 4.5$_{0.8}$ & \textbf{14.7}$_{2.2}$ & \multicolumn{1}{!{\vrule width 0.5pt}c}{} & 16.4 \\
 & SERA-8B \citep{shen2026serasoftverifiedefficientrepository} & 5.9$_{1.1}$ & 31.7\rlap{\textsuperscript{$\bullet$}}$_{1.7}$ & 31.5\rlap{\textsuperscript{$\bullet$}}$_{0.7}$ & 2.6\rlap{\textsuperscript{$\bullet$}}$_{0.8}$ & \multicolumn{1}{!{\vrule width 0.5pt}c}{} & \textbf{19.1}\rlap{\textsuperscript{$\bullet$}}$_{1.0}$ & 24.1$_{1.8}$ & 12.1$_{2.2}$ & 2.1$_{0.6}$ & 1.3$_{1.2}$ & \multicolumn{1}{!{\vrule width 0.5pt}c}{} & 13.3 \\
 & Qwen3-8B \citep{qwen3technicalreport} & 5.9$_{0.8}$ & 13.7\rlap{\textsuperscript{$\bullet$}}$_{1.4}$ & 13.2\rlap{\textsuperscript{$\bullet$}}$_{0.6}$ & 2.2\rlap{\textsuperscript{$\ast$}}$_{0.5}$ & \multicolumn{1}{!{\vrule width 0.5pt}c}{} & 11.7\rlap{\textsuperscript{$\bullet$}}$_{0.9}$ & 34.4$_{1.6}$ & 3.8$_{0.6}$ & 5.0$_{0.6}$ & 1.3$_{0.7}$ & \multicolumn{1}{!{\vrule width 0.5pt}c}{} & 10.2 \\
 & SETA-RL-Qwen3-8B \citep{seta2026camelai} & 6.4$_{1.0}$ & 0.7$_{0.5}$ & 1.4$_{0.3}$ & 1.5$_{0.7}$ & \multicolumn{1}{!{\vrule width 0.5pt}c}{} & 2.1$_{0.4}$ & 25.2$_{1.7}$ & 3.2$_{0.6}$ & 6.8$_{0.7}$ & 4.7$_{1.5}$ & \multicolumn{1}{!{\vrule width 0.5pt}c}{} & 6.4 \\
 & SWE-agent-LM-7B \citep{yang2025swesmithscalingdatasoftware} & 1.7\rlap{\textsuperscript{$\dagger$}}$_{0.3}$ & 15.3\rlap{\textsuperscript{$\dagger$}}$_{1.3}$ & 16.2\rlap{\textsuperscript{$\dagger$}}$_{0.6}$ & 0.7\rlap{\textsuperscript{$\dagger$}}$_{0.4}$ & \multicolumn{1}{!{\vrule width 0.5pt}c}{} & 3.1\rlap{\textsuperscript{$\dagger$}}$_{0.4}$ & 0.0\rlap{\textsuperscript{$\dagger$}}$_{0.0}$ & 0.8\rlap{\textsuperscript{$\dagger$}}$_{0.3}$ & 0.5$_{0.4}$ & 0.0$_{0.0}$ & \multicolumn{1}{!{\vrule width 0.5pt}c}{} & 3.0 \\
 & Llama-3.1-Nemotron-Nano-8B-v1 \citep{bercovich2025llamanemotronefficientreasoningmodels} & 1.0$_{0.5}$ & 0.0$_{0.0}$ & 0.4$_{0.0}$ & 0.0$_{0.0}$ & \multicolumn{1}{!{\vrule width 0.5pt}c}{} & 2.5$_{0.1}$ & 2.2$_{0.6}$ & 0.2$_{0.2}$ & 0.0$_{0.0}$ & 0.0$_{0.0}$ & \multicolumn{1}{!{\vrule width 0.5pt}c}{} & 0.8 \\
 & OpenThinker3-7B \citep{guha2025openthoughtsdatarecipesreasoning} & 1.0$_{0.0}$ & 0.0$_{0.0}$ & 0.4$_{0.0}$ & 0.7$_{0.5}$ & \multicolumn{1}{!{\vrule width 0.5pt}c}{} & 2.7$_{0.0}$ & 0.3$_{0.3}$ & 0.0$_{0.0}$ & 0.0$_{0.0}$ & 0.0$_{0.0}$ & \multicolumn{1}{!{\vrule width 0.5pt}c}{} & 0.6 \\
 & DeepSeek-R1-Distill-Qwen-7B \citep{guo_deepseek-r1_2025} & 1.0$_{0.0}$ & 0.0$_{0.0}$ & 0.4$_{0.0}$ & 0.4$_{0.4}$ & \multicolumn{1}{!{\vrule width 0.5pt}c}{} & 2.5$_{0.1}$ & 0.0$_{0.0}$ & 0.0$_{0.0}$ & 0.3$_{0.3}$ & 0.0$_{0.0}$ & \multicolumn{1}{!{\vrule width 0.5pt}c}{} & 0.5 \\
\midrule
\multirow{18}{*}{\rotatebox{90}{\textit{~32B Scale}}} & \textcolor{green!60!black}{OpenThinkerAgent-32B} & 41.3$_{2.1}$ & \textbf{55.7}$_{1.3}$ & 54.0$_{0.7}$ & \textbf{26.2}$_{1.6}$ & \multicolumn{1}{!{\vrule width 0.5pt}c}{} & \textbf{32.4}$_{1.2}$ & \textbf{85.9}$_{1.1}$ & 47.8$_{1.0}$ & \textbf{23.6}$_{1.5}$ & \textbf{44.0}$_{2.8}$ & \multicolumn{1}{!{\vrule width 0.5pt}c}{} & 44.8 \\
 & \textcolor{green!60!black}{Nemotron-Terminal-32B} \citep{pi2026dataengineeringscalingllm} & \textbf{48.6}$_{1.9}$ & 45.0$_{1.9}$ & 41.9$_{0.8}$ & \textbf{25.1}$_{1.5}$ & \multicolumn{1}{!{\vrule width 0.5pt}c}{} & 24.9$_{1.3}$ & 69.1$_{1.1}$ & \textbf{62.6}$_{0.9}$ & \textbf{22.3}$_{1.2}$ & 40.7$_{3.5}$ & \multicolumn{1}{!{\vrule width 0.5pt}c}{} & 40.9 \\
 & GLM-4.7-Flash \citep{5team2025glm45agenticreasoningcoding} & 24.5$_{1.9}$ & 46.7$_{1.6}$ & 45.0$_{0.8}$ & 15.4$_{1.5}$ & \multicolumn{1}{!{\vrule width 0.5pt}c}{} & 20.1$_{1.1}$ & 77.8$_{1.3}$ & 40.9$_{1.1}$ & \textbf{22.8}$_{1.7}$ & \textbf{42.0}$_{3.0}$ & \multicolumn{1}{!{\vrule width 0.5pt}c}{} & 37.7 \\
 & Nemotron-Terminal-14B \citep{pi2026dataengineeringscalingllm} & 34.7$_{1.9}$ & 35.3$_{1.8}$ & 31.5$_{0.8}$ & 21.0$_{1.6}$ & \multicolumn{1}{!{\vrule width 0.5pt}c}{} & 18.4$_{1.2}$ & 75.9$_{1.5}$ & 56.9$_{1.1}$ & 16.3$_{1.2}$ & 30.7$_{3.3}$ & \multicolumn{1}{!{\vrule width 0.5pt}c}{} & 35.8 \\
 & \textcolor{green!60!black}{SWE-Lego-Qwen3-32B} \citep{tao2026swelego} & 25.6\rlap{\textsuperscript{$\ast$}}$_{1.9}$ & 51.0\rlap{\textsuperscript{$\ast$}}$_{1.5}$ & 51.0\rlap{\textsuperscript{$\ast$}}$_{0.7}$ & 16.1\rlap{\textsuperscript{$\ast$}}$_{1.5}$ & \multicolumn{1}{!{\vrule width 0.5pt}c}{} & 30.1\rlap{\textsuperscript{$\ast$}}$_{1.1}$ & 81.0\rlap{\textsuperscript{$\ast$}}$_{0.8}$ & 36.2$_{1.1}$ & 12.9\rlap{\textsuperscript{$\ast$}}$_{1.2}$ & 15.3$_{2.4}$ & \multicolumn{1}{!{\vrule width 0.5pt}c}{} & 34.7 \\
 & daVinci-Dev-32B \citep{zeng2026davincidevagentnativemidtrainingsoftware} & 22.2$_{1.9}$ & \textbf{55.3}\rlap{\textsuperscript{$\dagger$}}$_{1.3}$ & 54.5\rlap{\textsuperscript{$\dagger$}}$_{0.8}$ & 12.0$_{1.6}$ & \multicolumn{1}{!{\vrule width 0.5pt}c}{} & 21.3\rlap{\textsuperscript{$\dagger$}}$_{0.9}$ & 64.8$_{1.8}$ & 38.0$_{1.2}$ & 11.5$_{1.4}$ & 21.3$_{3.0}$ & \multicolumn{1}{!{\vrule width 0.5pt}c}{} & 31.9 \\
 & Qwen3-Coder-30B-A3B-Instruct \citep{qwen3technicalreport} & 23.3\rlap{\textsuperscript{$\ast$}}$_{2.0}$ & 47.7\rlap{\textsuperscript{$\ast$}}$_{1.5}$ & \underline{51.6}\rlap{\textsuperscript{$\ast$}}$_{0.0}$ & 13.9\rlap{\textsuperscript{$\ast$}}$_{1.4}$ & \multicolumn{1}{!{\vrule width 0.5pt}c}{} & 27.6\rlap{\textsuperscript{$\ast$}}$_{1.1}$ & 67.8$_{1.3}$ & 20.9$_{1.0}$ & 11.3\rlap{\textsuperscript{$\ast$}}$_{1.0}$ & 12.7$_{2.0}$ & \multicolumn{1}{!{\vrule width 0.5pt}c}{} & 29.4 \\
 & \textcolor{green!60!black}{SERA-32B} \citep{shen2026serasoftverifiedefficientrepository} & 20.9$_{1.7}$ & 48.7\rlap{\textsuperscript{$\bullet$}}$_{1.6}$ & 49.4\rlap{\textsuperscript{$\bullet$}}$_{0.7}$ & 9.7\rlap{\textsuperscript{$\bullet$}}$_{1.2}$ & \multicolumn{1}{!{\vrule width 0.5pt}c}{} & 26.7\rlap{\textsuperscript{$\bullet$}}$_{1.1}$ & 69.1$_{1.6}$ & 15.6$_{1.0}$ & 8.7$_{1.2}$ & 17.3$_{2.6}$ & \multicolumn{1}{!{\vrule width 0.5pt}c}{} & 28.1 \\
 & \textcolor{green!60!black}{SA-SWE-32B} \citep{cao2025skyrlagentefficientrltraining} & 15.3$_{1.8}$ & 36.0\rlap{\textsuperscript{$\P$}}$_{0.0}$ & \underline{39.4}\rlap{\textsuperscript{$\P$}}$_{0.0}$ & \underline{16.2}\rlap{\textsuperscript{$\ast$}}$_{0.0}$ & \multicolumn{1}{!{\vrule width 0.5pt}c}{} & 17.3\rlap{\textsuperscript{$\ast$}}$_{1.0}$ & 74.8$_{1.6}$ & 15.8\rlap{\textsuperscript{$\ast$}}$_{1.0}$ & 11.5$_{0.9}$ & 13.3$_{2.5}$ & \multicolumn{1}{!{\vrule width 0.5pt}c}{} & 26.9 \\
 & \textcolor{green!60!black}{DeepSWE-Preview} \citep{deepswe2025} & 13.3$_{1.6}$ & 38.7\rlap{\textsuperscript{$\S$}}$_{0.0}$ & \underline{42.2}\rlap{\textsuperscript{$\S$}}$_{0.0}$ & 4.9$_{1.0}$ & \multicolumn{1}{!{\vrule width 0.5pt}c}{} & 27.3\rlap{\textsuperscript{$\dagger$}}$_{1.3}$ & 77.2$_{1.4}$ & 8.7$_{0.8}$ & 16.5$_{1.4}$ & 10.0$_{2.2}$ & \multicolumn{1}{!{\vrule width 0.5pt}c}{} & 26.7 \\
 & Nemotron-Nano-30B-A3B \citep{nvidia2025nemotron3nano} & 22.4$_{2.1}$ & 33.3\rlap{\textsuperscript{$\ast$}}$_{1.7}$ & 30.6\rlap{\textsuperscript{$\ast$}}$_{0.7}$ & 6.4\rlap{\textsuperscript{$\ast$}}$_{1.2}$ & \multicolumn{1}{!{\vrule width 0.5pt}c}{} & 21.5\rlap{\textsuperscript{$\ast$}}$_{1.1}$ & 81.6\rlap{\textsuperscript{$\ast$}}$_{1.5}$ & 16.2$_{1.0}$ & 14.7\rlap{\textsuperscript{$\ast$}}$_{1.1}$ & 11.3$_{2.6}$ & \multicolumn{1}{!{\vrule width 0.5pt}c}{} & 26.0 \\
 & SWE-agent-LM-32B \citep{yang2025swesmithscalingdatasoftware} & 22.3$_{1.8}$ & 41.7\rlap{\textsuperscript{$\dagger$}}$_{1.2}$ & 40.7\rlap{\textsuperscript{$\dagger$}}$_{0.6}$ & 13.9$_{1.4}$ & \multicolumn{1}{!{\vrule width 0.5pt}c}{} & 20.0\rlap{\textsuperscript{$\dagger$}}$_{0.9}$ & 40.9$_{2.1}$ & 32.9$_{1.2}$ & 8.7$_{1.0}$ & 11.3$_{2.6}$ & \multicolumn{1}{!{\vrule width 0.5pt}c}{} & 24.1 \\
 & \textcolor{green!60!black}{Qwen3-32B} \citep{qwen3technicalreport} & 13.7$_{1.8}$ & 26.7\rlap{\textsuperscript{$\bullet$}}$_{1.3}$ & 29.1\rlap{\textsuperscript{$\bullet$}}$_{0.7}$ & 7.5\rlap{\textsuperscript{$\ast$}}$_{0.9}$ & \multicolumn{1}{!{\vrule width 0.5pt}c}{} & 28.9\rlap{\textsuperscript{$\bullet$}}$_{1.2}$ & 68.3$_{1.4}$ & 6.8$_{0.7}$ & 9.7$_{1.0}$ & 9.3$_{1.8}$ & \multicolumn{1}{!{\vrule width 0.5pt}c}{} & 22.8 \\
 & LiteCoder-Terminal-30B-A3B \citep{litecoder2025blog} & 29.7$_{2.2}$ & 19.3$_{1.6}$ & 16.7$_{0.7}$ & 13.5$_{1.5}$ & \multicolumn{1}{!{\vrule width 0.5pt}c}{} & 22.2$_{1.1}$ & 55.0$_{1.6}$ & 23.1$_{1.1}$ & 11.5$_{1.1}$ & 16.0$_{2.7}$ & \multicolumn{1}{!{\vrule width 0.5pt}c}{} & 22.6 \\
 & Qwen2.5-Coder-32B-Instruct \citep{hui2024qwen25codertechnicalreport} & 10.1\rlap{\textsuperscript{$\ast$}}$_{1.4}$ & 14.7\rlap{\textsuperscript{$\ast$}}$_{1.4}$ & 10.0\rlap{\textsuperscript{$\ast$}}$_{0.5}$ & 7.1\rlap{\textsuperscript{$\ast$}}$_{1.0}$ & \multicolumn{1}{!{\vrule width 0.5pt}c}{} & 13.3\rlap{\textsuperscript{$\ast$}}$_{0.8}$ & 60.4\rlap{\textsuperscript{$\ast$}}$_{0.9}$ & 37.4$_{1.1}$ & 6.8$_{1.0}$ & 14.7$_{2.3}$ & \multicolumn{1}{!{\vrule width 0.5pt}c}{} & 21.4 \\
 & R2EGym-32B-Agent \citep{jain2025r2egymproceduralenvironmentshybrid} & 14.9$_{1.7}$ & 33.0\rlap{\textsuperscript{$\S$}}$_{0.0}$ & \underline{34.4}\rlap{\textsuperscript{$\S$}}$_{0.0}$ & 9.7\rlap{\textsuperscript{$\ast$}}$_{0.9}$ & \multicolumn{1}{!{\vrule width 0.5pt}c}{} & 15.9\rlap{\textsuperscript{$\ast$}}$_{0.7}$ & 49.1$_{1.9}$ & 25.2$_{1.2}$ & 6.0$_{0.9}$ & 2.7$_{1.3}$ & \multicolumn{1}{!{\vrule width 0.5pt}c}{} & 20.4 \\
 & Qwen3-30B-A3B-Instruct-2507 \citep{qwen3technicalreport} & 15.6$_{1.6}$ & 5.7$_{1.1}$ & 6.7$_{0.5}$ & 6.7$_{1.2}$ & \multicolumn{1}{!{\vrule width 0.5pt}c}{} & 19.9$_{1.1}$ & 58.5$_{1.4}$ & 12.1$_{0.8}$ & 8.1$_{0.9}$ & 6.7$_{1.8}$ & \multicolumn{1}{!{\vrule width 0.5pt}c}{} & 17.0 \\
 & \textcolor{orange}{OpenSWE-32B} \citep{fu2026davincienvopensweenvironment} & 4.7\rlap{\textsuperscript{$\dagger$}}$_{0.9}$ & \textcolor{orange}{44.0\rlap{\textsuperscript{$\dagger$}}$_{1.9}$} & \textcolor{orange}{\underline{\textbf{62.4}}\rlap{\textsuperscript{$\dagger$}}$_{0.0}$} & 3.4\rlap{\textsuperscript{$\dagger$}}$_{0.8}$ & \multicolumn{1}{!{\vrule width 0.5pt}c}{} & 10.1\rlap{\textsuperscript{$\dagger$}}$_{0.9}$ & 0.5$_{0.4}$ & 3.0\rlap{\textsuperscript{$\dagger$}}$_{0.5}$ & 0.0\rlap{\textsuperscript{$\dagger$}}$_{0.0}$ & 0.0$_{0.0}$ & \multicolumn{1}{!{\vrule width 0.5pt}c}{} & 11.3 \\
\midrule
\multirow{7}{*}{\rotatebox{90}{\textit{Frontier}}} & Kimi-K2.5 \citep{kimiteam2026kimik25visualagentic} & \textbf{72.1}$_{1.9}$ & \textbf{70.3}$_{1.3}$ & 70.7$_{0.6}$ & 40.1$_{1.8}$ & \multicolumn{1}{!{\vrule width 0.5pt}c}{} & \textbf{71.9}$_{1.1}$ & \textbf{86.4}$_{0.9}$ & 49.4$_{0.9}$ & \textbf{58.3}$_{1.7}$ & 65.3$_{3.1}$ & \multicolumn{1}{!{\vrule width 0.5pt}c}{} & 63.2 \\
 & GLM-5 \citep{glm5team2026vibecoding} & \textbf{70.6}$_{2.0}$ & \textbf{70.0}$_{1.4}$ & \underline{\textbf{72.8}}\rlap{\textsuperscript{$\star$}}$_{0.0}$ & \textbf{46.1}$_{1.7}$ & \multicolumn{1}{!{\vrule width 0.5pt}c}{} & 56.3$_{1.1}$ & \textbf{86.2}$_{0.7}$ & 45.3$_{0.7}$ & \textbf{58.3}$_{1.7}$ & \textbf{72.0}$_{2.6}$ & \multicolumn{1}{!{\vrule width 0.5pt}c}{} & 62.4 \\
 & Qwen3.5-27B \citep{qwen3.5} & 62.2$_{2.0}$ & 60.3$_{1.3}$ & 62.7$_{0.6}$ & 40.1$_{1.8}$ & \multicolumn{1}{!{\vrule width 0.5pt}c}{} & 50.5$_{1.0}$ & \textbf{86.4}$_{0.8}$ & \textbf{64.7}$_{0.7}$ & 44.9$_{1.6}$ & 55.3$_{2.8}$ & \multicolumn{1}{!{\vrule width 0.5pt}c}{} & 57.8 \\
 & GLM-4.7-FP8 \citep{5team2025glm45agenticreasoningcoding} & 61.2$_{2.1}$ & 64.7\rlap{\textsuperscript{$\ast$}}$_{1.7}$ & 63.9\rlap{\textsuperscript{$\ast$}}$_{0.7}$ & 32.7$_{1.4}$ & \multicolumn{1}{!{\vrule width 0.5pt}c}{} & 46.5$_{1.1}$ & \textbf{86.2}$_{0.6}$ & 55.4$_{0.8}$ & 45.4$_{1.7}$ & 60.0$_{3.1}$ & \multicolumn{1}{!{\vrule width 0.5pt}c}{} & 55.7 \\
 & Qwen3.5-9B \citep{qwen3.5} & 32.7$_{2.0}$ & 48.7$_{1.8}$ & 49.0$_{0.8}$ & 19.9$_{1.7}$ & \multicolumn{1}{!{\vrule width 0.5pt}c}{} & 30.1$_{1.1}$ & 79.4$_{1.1}$ & 47.6$_{1.2}$ & 24.9$_{1.6}$ & 48.0$_{2.8}$ & \multicolumn{1}{!{\vrule width 0.5pt}c}{} & 42.7 \\
 & Qwen3-Coder-480B-A35B-Instruct-FP8 \citep{qwen3technicalreport} & 41.1$_{2.2}$ & 44.0$_{1.8}$ & 40.2$_{0.8}$ & 21.0$_{1.7}$ & \multicolumn{1}{!{\vrule width 0.5pt}c}{} & 44.3$_{1.2}$ & 73.4$_{1.1}$ & 36.2$_{0.8}$ & 22.6$_{1.5}$ & 38.7$_{2.6}$ & \multicolumn{1}{!{\vrule width 0.5pt}c}{} & 39.5 \\
 & Qwen3-235B-A22B-Instruct \citep{qwen3technicalreport} & 40.5$_{2.2}$ & 15.7$_{1.6}$ & 15.5$_{0.7}$ & 13.1$_{1.7}$ & \multicolumn{1}{!{\vrule width 0.5pt}c}{} & 34.8$_{1.2}$ & 81.6$_{1.0}$ & 46.2$_{0.9}$ & 15.5$_{1.4}$ & 27.3$_{2.8}$ & \multicolumn{1}{!{\vrule width 0.5pt}c}{} & 33.4 \\
\bottomrule
\end{tabular}}
\vspace{6pt}
\caption{\textbf{OpenThinkerAgent-32B is the strongest 32B-Scale model by average accuracy across the seven benchmarks shared with main Table~\ref{tab:ot_agent_main_table1}.} Each cell reports the model's best accuracy (\%) across evaluated agent harnesses; the subscript is that cell's per-run standard error. The Avg column averages each model's accuracy over the same seven benchmarks used in main Table~\ref{tab:ot_agent_main_table1} (SWE-Bench-Verified, Terminal-Bench-2, Aider-Polyglot, BFCL-Parity, MedAgentBench, GAIA-127, FinanceAgent-Terminal); OpenThoughts-TBLite and SWE-Bench-Verified-100 are shown as columns but excluded from the average. Models are grouped into scale bands (8B, 32B, Frontier). Within each band, cells whose accuracy lies within 1 Standard Error of the group's maximum on that benchmark are bolded.\\[\baselineskip]\textbf{Harness symbols.} Cells without a superscript use the default harness terminus-2. Non-default symbols: $\ast$ = OpenHands \citep{wang2025openhands}, $\dagger$ = regular SWE-Agent \citep{yang2024sweagent}, $\bullet$ = SERA-SWE-Agent (the SERA team's SWE-Agent config), $\S$ = R2EGym-Edit-Agent (a custom ReAct scaffold implemented in R2EGym evaluation \citep{jain2025r2egymproceduralenvironmentshybrid}), $\P$ = SkyRL (a simple ReAct \citep{yao2023react} scaffold used in SkyRL-Agent evaluation \citep{cao2025skyrl}), $\star$ = Mini-SWE-Agent \citep{yang2024sweagent}. Note that R2EGym-Edit-Agent and SkyRL-Agent are not directly available within Harbor's agent suite.\\[\baselineskip]\textbf{Color coding.} \textcolor{green!60!black}{Green} model names also appear in main Table~\ref{tab:ot_agent_main_table1}. \textcolor{orange}{Orange} flags models whose published results we have not been able to fully reproduce (see \cref{app:eval-setup} for more detail). Their SWE-Bench-Verified and SWE-Bench-Verified-100 cells are also rendered in orange to flag the gap. \underline{Underlined} numbers are filled in from the model's reported paper number rather than our reproduction. Note: we do not include in this table open-data releases that did not release a model, such as CoderForge-Preview~\citep{CoderForge2026}.}
\label{tab:appendix_eval_table}
\end{table}

\paragraph{Evaluation on Terminal-Bench 2.1.}
Concurrent with our experiments, Terminal-Bench 2.1 was released, which revises 31\% of Terminal-Bench 2.0 tasks (28/89 tasks) to correct for three defect classes: specification--verification mismatches, resource mismatches, and benchmark drift ~\citep{tb21audit}. Because Terminal-Bench 2.0 is one of our two core benchmarks, we re-evaluated a representative baseline (Qwen 3.5-27B) on Terminal-Bench 2.1 to verify that our conclusions hold after a new release.  The table \Cref{tab:tb21} reports the accuracy in TB2.0 and TB2.1.  The observed change (~3.0 pp) is far smaller than the up-to-12.0 pp shifts the audit reports for frontier agents~\citep{tb21audit}. We therefore continue to report TB2.0 results throughout, as re-evaluating every model on TB2.1 would be cost-prohibitive.

\begin{table}[h]
\centering
\caption{Qwen 3.5-27B improves slightly on Terminal-Bench 2.1; however, the gap between 2.1 and 2.0 is far smaller than the observed gap for frontier agents. This helps motivate our decision to report Terminal-Bench 2.0 results in the paper.}
\begin{tabular}{lc}
\toprule
Benchmark & Accuracy $\pm$ se (\%) \\
\midrule
Terminal-Bench 2.0 & $40.1 \pm 1.6$ \\
Terminal-Bench 2.1 & $43.1 \pm 1.4$ \\
\bottomrule
\end{tabular}
\label{tab:tb21}
\end{table}



\end{document}